\documentclass{ieeeaccess}
\usepackage{cite}
\usepackage{amsmath,amssymb,amsfonts}
\usepackage{algorithmic}
\usepackage{graphicx}
\usepackage{textcomp}
\usepackage{url}
\usepackage{gensymb}
\def\BibTeX{{\rm B\kern-.05em{\sc i\kern-.025em b}\kern-.08em
    T\kern-.1667em\lower.7ex\hbox{E}\kern-.125emX}}
\begin{document}
\history{Date of publication xxxx 00, 0000, date of current version xxxx 00, 0000.}
\doi{10.1109/ACCESS.2023.0322000}

\title{Archangel: A Hybrid UAV-based Human Detection Benchmark with Position and Pose Metadata}
\author{\uppercase{Yi-Ting~Shen}\authorrefmark{1},
\uppercase{Yaesop~Lee}\authorrefmark{1},
\uppercase{Heesung~Kwon}\authorrefmark{2}, 
\uppercase{Damon~M.~Conover}\authorrefmark{2}, 
\uppercase{Shuvra~S.~Bhattacharyya}\authorrefmark{1}, 
\uppercase{Nikolas~Vale}\authorrefmark{2},
\uppercase{Joshua~D.~Gray}\authorrefmark{3},
\uppercase{G.~Jeremy~Leong}\authorrefmark{4},
\uppercase{Kenneth~Evensen}\authorrefmark{5},
\uppercase{and Frank~Skirlo}\authorrefmark{5}}

\address[1]{University of Maryland, ECE Department and UMIACS, College Park, MD, USA}
\address[2]{DEVCOM Army Research Laboratory (ARL), Adelphi, MD, USA}
\address[3]{Fibertek Inc., Herndon, VA, USA}
\address[4]{Department of Energy, Washington, DC, USA}
\address[5]{Defense Threat Reduction Agency (DTRA), Fort Belvoir, VA, USA}
\tfootnote{This research was sponsored by the Defense Threat Reduction Agency (DTRA) and the DEVCOM Army Research Laboratory (ARL) under Grant No. W911NF2120076.}

\markboth
{Shen \headeretal: Preparation of Papers for IEEE TRANSACTIONS and JOURNALS}
{Shen \headeretal: Preparation of Papers for IEEE TRANSACTIONS and JOURNALS}

\corresp{Corresponding author: Yi-Ting~Shen (e-mail: ytshen@umd.edu).}

\begin{abstract}
Learning to detect objects, such as humans, in imagery captured by an unmanned aerial vehicle (UAV) usually suffers from tremendous variations caused by the UAV's position towards the objects. In addition, existing UAV-based benchmark datasets do not provide adequate dataset metadata, which is essential for precise model diagnosis and learning features invariant to those variations. In this paper, we introduce \textit{Archangel}, the first UAV-based object detection dataset composed of real and synthetic subsets captured with similar imagining conditions and UAV position and object pose metadata. A series of experiments are carefully designed with a state-of-the-art object detector to demonstrate the benefits of leveraging the metadata during model evaluation. Moreover, several crucial insights involving both real and synthetic data during model optimization are presented. In the end, we discuss the advantages, limitations, and future directions regarding \textit{Archangel} to highlight its distinct value for the broader machine learning community.
\end{abstract}

\begin{keywords}
UAV-based object detection, human detection, UAV-based benchmark dataset, position metadata, synthetic data, model optimization.
\end{keywords}

\titlepgskip=-21pt

\maketitle

\section{Introduction}
\IEEEPARstart{W}{ith}
the recent rapid advancement in edge computing technology coupled with resource-constrained mobile platforms, particularly unmanned aerial vehicles (UAVs) with electro-optical (EO) sensor payloads, a wide range of UAV-enabled applications have been more prevalent. Notable examples include UAV-enabled search and rescue in disaster management \cite{erdelj2016uav}, aerial surveillance and reconnaissance for civilian and military purposes \cite{iscc2020}, precision agriculture \cite{agri}, traffic analysis \cite{traffic}, and intelligent transportation applications \cite{intelli}. Lately, owing to the remarkable progress of artificial intelligence and machine learning technology, tailored to distinct constraints of small UAV platforms, these UAV-based applications have been frequently providing promising solutions and successfully achieving the goals and operational requirements of their corresponding applications. Central to the above UAV-based applications are streamlined plug-ins that can effectively interrogate real-time imagery, captured with UAVs, and provide image/video analytics relevant to UAV-based scene understanding, particularly object detection and recognition.

Recently, extensive efforts in object detection and recognition have led to extraordinary advances in the perception accuracy in various challenges associated with large-scale object detection benchmarks that were captured primarily with ground-based cameras \cite{imagenet, pascal, coco}. Compared to ground-based object detection and recognition, UAV-based object detection poses unique and severe challenges, as UAV flight inevitably results in a wider range of variations in the conditions for capturing images, including the altitudes and viewing angles of cameras, system turbulence, and weather events. These variations lead to more drastic variations in object appearances/attributes, and thus pose additional challenges to onboard detection models in the location and recognition of objects of interest. In general, changes in object appearances/attributes, caused by varying the image collection conditions, entail three major dependencies: (1) pose dependency caused by changes in the UAV position or camera viewing angle, (2) scale dependency owing to the distance between the UAV and object, and (3) image quality dependency due to UAV turbulence or various weather conditions.

We argue that developing object detection models which can adequately learn features invariant to these dependencies is key to substantially enhancing object detection accuracy in UAV-based perception. This requires curating UAV-based datasets that include images and metadata that carefully depict the full spectrum of correlations between the target scene on the ground and the camera on the UAV, whose imaging conditions constantly change as the UAV navigates the entire range of given operational requirements. Therefore, it is imperative to have a UAV-based object detection benchmark carefully curated with object poses, UAV positions, and weather information in the form of metadata for accurate model validation and verification. 

Existing UAV-based benchmarks, such as VisDrone \cite{visdrone}, UAVDT \cite{UAVDT}, Okutama-Action \cite{Oku}, and Standford Drone Dataset \cite{stanford}, provide limited metadata. Despite containing a wide variety of scenes captured using UAVs under different circumstances with various types of objects of interest, they do not provide a complete set of metadata, such as object poses and UAV positions, for each image in the datasets. This significant lack of information about how objects on the ground are projected through the camera lens as a function of UAV positions can lead to considerable limitations in learning about the objects in UAV-based images, as the appearances/attributes of the objects are subject to large variations.

\begin{figure*}
\centering
\includegraphics[width=\textwidth]{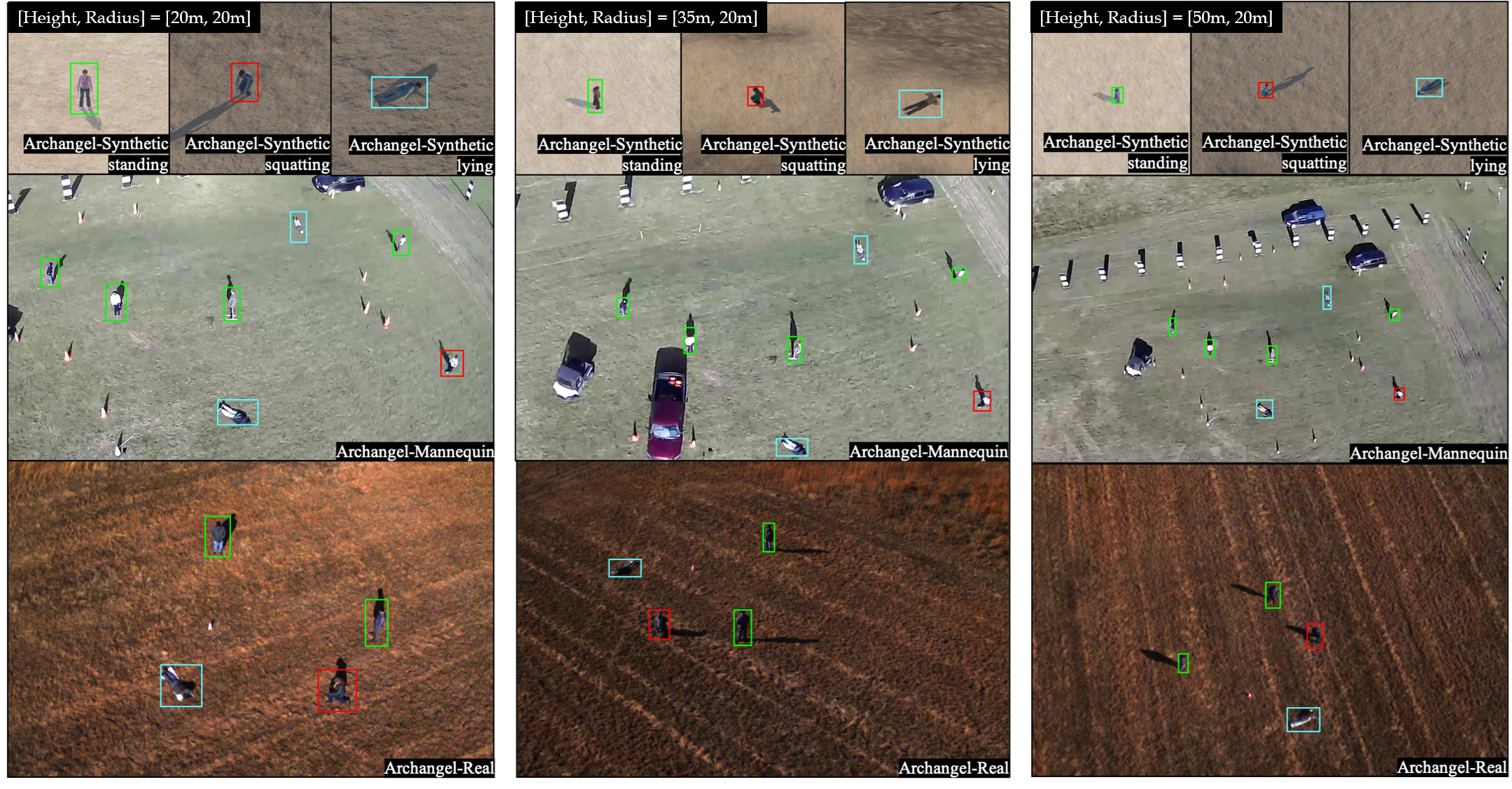}
\caption{Examples of images in \textit{Archangel-Synthetic} (top), \textit{Archangel-Mannequin} (middle), and \textit{Archangel-Real} (bottom). Each image is labeled with its UAV position [Height, Radius]: [20m, 20m] (left), [35m, 20m] (middle), and [50m, 20m] (right). Each instance is labelled with its object pose: \textit{stand} (\textcolor{green}{green}), \textit{squat} or \textit{kneel} (\textcolor{red}{red}), and \textit{prone} (\textcolor{cyan}{cyan}).}
\label{fig:dataset}
\end{figure*}

\begin{figure}
\centering
\includegraphics[width=0.8\columnwidth]{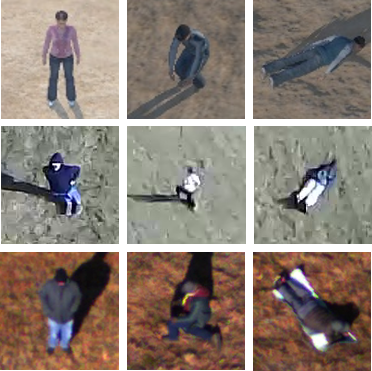}
\caption{Examples of the three poses in \textit{Archangel}: \textit{stand} (left), \textit{squat} or \textit{kneel} (middle), and \textit{prone} (right).}
\label{fig:pose}
\end{figure}

To overcome the limitation resulting from lack of metadata, we introduce a large-scale UAV-based dataset, called \textit{Archangel}, collected with comprehensive position and pose information by the DEVCOM Army Research Laboratory (ARL) (Fig.~\ref{fig:dataset}). \textit{Archangel} comprises three sub-datasets: \textit{Archangel-Real} \cite{realarchangel}, \textit{Archangel-Mannequin} \cite{realarchangel} and \textit{Archangel-Synthetic} \cite{syn}. \textit{Archangel-Real} comprises video sequences captured by a UAV flying at various altitudes and radii of rotation circles. It sets a group of real humans as targets, and each human is in one of three possible poses (i.e., \textit{stand}, \textit{kneel} or \textit{squat}\footnote{The terms \textit{kneel} and \textit{squat} are used interchangeably throughout this paper.}, and \textit{prone}) (Fig.~\ref{fig:pose}). Similarly, \textit{Archangel-Mannequin} sets a group of mannequins and different types of vehicles as targets. The imaging conditions for these two sub-datasets, such as UAV altitudes, ranges to targets, and object poses, are the same. Unlike \textit{Archangel-Real} and \textit{Archangel-Mannequin}, which were collected in real-world environments, \textit{Archangel-Synthetic} was generated using the Unity game engine \cite{unity}. It includes a number of different virtual characters, who are in the same poses as described above and rendered with diverse illumination conditions. \textit{Archangel-Synthetic} is designed for augmenting the other two real sub-datasets and for studying various issues tied to optimizing machine learning (ML) models using synthetic data, such as synthetic data augmentation and domain adaptation \cite{nikolenko2021synthetic}, with respect to UAV-based scene understanding.

In addition to collecting a new dataset, we further characterize \textit{Archangel} using state-of-the-art (SoTA) object detection models, specifically the YOLOv5 family \cite{yolo} which has five different levels of architectural complexity and is pre-trained on MS-COCO \cite{coco}, a large-scale ground-based object detection dataset. In this paper, we focus on the YOLOv5 models with lower complexity (i.e., $YOLOv5n6$, $YOLOv5s6$, $YOLOv5m6$) since they are able to run on computing resource typically available on small UAV platforms (Tab. \ref{tab:yolo}). For each model, we evaluate its human detection performance using \textit{Archangel-Real}. Since we programmed our UAVs to circle around the human targets at various altitudes and radii during data collection (Fig.~\ref{fig:collection}), the detection accuracy can be compared across the whole range of UAV positions.

Furthermore, we optimize the pre-trained YOLOv5 models with different fine-tuning strategies using various hybrid combinations of subsets from \textit{Archangel-Mannequin} and \textit{Archangel-Synthetic}. These fine-tuning strategies have been designed to provide valuable guidelines for leveraging synthetic data in training ML models to boost their performance. A comprehensive performance comparison between the baseline and the optimized YOLOv5 models is presented to demonstrate how incorporating a combination of real and synthetic data can enhance detection performance across varying UAV positions (Sec. \ref{sec:results} and \ref{sec:ablation}). One of the critical findings from the comparative performance analysis indicates that if the real and synthetic data used for fine-tuning is balanced with respect to the amount of data, a significant performance boost can be achieved even with a low-complexity model, such as $YOLOv5n6$. Furthermore, the optimization based on the real and synthetic data is much more effective on infrequent object poses that are rarely seen in the original dataset for pre-training. For example, \textit{prone} from \textit{Archangel} is not often seen in MS-COCO \cite{coco}, so the performance improvement on this pose is more evident.

This paper is extended from our previous preliminary studies \cite{syn, realarchangel}, where we mainly focused on introducing new datasets separately without extensive data analysis and characterization. As a result, the major scope of this paper is to extensively study the three sub-datasets jointly as a unified UAV-based benchmark with metadata for human detection. In summary, the contributions of this paper are as follows:
\begin{enumerate}
   \item We present a unified \textit{Archangel} dataset\footnote{\textit{Archangel} is available for access through: \textit{https://a2i2-archangel.vision}} after substantially restructuring the three sub-datasets since our conference publications \cite{realarchangel, syn}, including additional labeling \textit{Archangel-Real} and an expansion of the range of \textit{Archangel-Synthetic}. Note that \textit{Archangel} was not previously available for access due to incomplete labeling and restructuring. To the best of our knowledge, \textit{Archangel} is the first UAV-based object detection dataset which contains real and synthetic sub-datasets captured with similar imaging conditions and includes an extensive set of metadata (e.g., object poses and UAV positions).   
   \item We conduct extensive data analysis on \textit{Archangel} by jointly analyzing its three sub-datasets (\textit{Archangel-Synthetic}, \textit{Archangel-Mannequin} and \textit{Archangel-Real}). In particular, we provide several important guidelines on exploiting real and synthetic data together to improve UAV-based object detectors.
\end{enumerate}

\section{Related Work}
\subsection{UAV-based Object Detection Datasets}

There is an increasing number of large-scale benchmarks for object detection, utilizing images captured with some fixed or moving cameras on the ground \cite{coco, pascal, OpenImages, sun2020scalability, dollar2009pedestrian, geiger2012we, shao2019objects365}, yet relatively few datasets have been collected with UAVs. Moreover, these UAV-based object detection datasets all have their own limitations. For instance, VisDrone \cite{visdrone} is one of the major datasets for UAV-based object detection. It consists of images captured with UAVs in dozens of different scenarios, in which ten categories of objects were selected and carefully labelled. While it does have an advantage in data diversity, VisDrone does not provide any metadata, such as UAV positions. In contrast to VisDrone, UAVDT \cite{UAVDT}, another well-known benchmark for UAV-based detection and tracking, provides three different kinds of UAV-specific metadata (i.e., weather condition, flying altitude and viewing angle) and a few object attributes such as vehicle category. However, the annotations for the metadata are coarse (i.e., 3 categories for each). Also, the dataset does not contain the human category. Unlike UAVDT, Okutama-Action \cite{Oku} is composed of images from aerial views and contains humans in different human poses. However, it provides limited metadata regarding UAV altitudes and camera viewing angles. A212-Haze \cite{narayanan2022multi}, which is the first real haze and object detection dataset with in-situ smoke measurements aligned to aerial imagery and is used in the recent $UG^2$+ Challenge \cite{ug2}, provides UAV position metadata. However, A212-Haze does not include a synthetic subset, limiting its usage for facilitating studies on how to improve UAV-based object detectors by using synthetic data. More recently, DGTA \cite{kiefer2022leveraging} generates synthetic datasets associated with existing UAV-based object detection datasets and provides UAV position metadata for the generated datasets. Nevertheless, the existing UAV-based datasets they use, such as VisDrone \cite{visdrone}, are still lack of metadata, limiting DGTA's usage for precise model diagnosis.

There are two other types of datasets which are closely related to UAV-based object detection. First, datasets such as DOTA \cite{ding2021object} include aerial images collected from satellites or aircraft. Although they are usually curated for remote sensing applications, detecting objects in such datasets also suffers from severe variations in the scale and orientation of the objects. Second, some UAV-based datasets are designed for certain vision tasks strongly associated with object detection. For instance, CARPK \cite{hsieh2017drone} is a large-scale car parking lot dataset designed for object counting. MOR-UAV \cite{MOR-UAV} is a large-scale moving object recognition dataset comprising videos captured by a UAV in various environments, such as urban areas and highways. Stanford Drone Dataset \cite{stanford} is used for analyzing various object trajectories in the real world from the top-view. UAV123 \cite{uav-eccv} is used for low altitude UAV-based object tracking. Similar to \textit{Archangel}, UAV123 also includes synthetic data generated by a photo-realistic simulator.

A comprehensive investigation of recent UAV-based datasets is shown in Tab.~\ref{tab:dataset-summary}. Note that \textit{Archangel} is the first ever UAV-based dataset collection not only containing both real and synthetic data but also providing an extensive set of metadata, including object poses and UAV positions.

\begin{table*}
\centering
\caption{Comparison of recent UAV-based datasets. (1k = 1000)}
\label{tab:dataset-summary}
\resizebox{\textwidth}{!}{%
\begin{tabular}{|cccccccc|ccc|}
\hline
\multicolumn{8}{|c|}{General Information} & \multicolumn{3}{c|}{Metadata} \\ \hline
\multicolumn{1}{|c|}{Name} & \multicolumn{1}{c|}{Tasks} & \multicolumn{1}{c|}{Year} & \multicolumn{1}{c|}{\#Clips} & \multicolumn{1}{c|}{\#Images} & \multicolumn{1}{c|}{Resolution} & \multicolumn{1}{c|}{Syn/Real} & Human & \multicolumn{1}{c|}{Obj. Poses} & \multicolumn{1}{c|}{UAV Pos.} & Lighting Cond. \\ \hline
\multicolumn{1}{|c|}{Stanford Drone Dataset \cite{stanford}} & \multicolumn{1}{c|}{TF} & \multicolumn{1}{c|}{2016} & \multicolumn{1}{c|}{60} & \multicolumn{1}{c|}{929.5k} & \multicolumn{1}{c|}{1400$\times$1904} & \multicolumn{1}{c|}{R} & \checkmark & \multicolumn{1}{c|}{-} & \multicolumn{1}{c|}{-} & - \\
\multicolumn{1}{|c|}{UAV123 \cite{uav-eccv}} & \multicolumn{1}{c|}{OT} & \multicolumn{1}{c|}{2016} & \multicolumn{1}{c|}{123} & \multicolumn{1}{c|}{112.6k} & \multicolumn{1}{c|}{1280$\times$720} & \multicolumn{1}{c|}{S+R} & \checkmark & \multicolumn{1}{c|}{-} & \multicolumn{1}{c|}{-} & - \\
\multicolumn{1}{|c|}{Okutama-Action \cite{Oku}} & \multicolumn{1}{c|}{OD, AR} & \multicolumn{1}{c|}{2017} & \multicolumn{1}{c|}{43} & \multicolumn{1}{c|}{77.4k} & \multicolumn{1}{c|}{3840$\times$2160} & \multicolumn{1}{c|}{R} & \checkmark & \multicolumn{1}{c|}{\checkmark} & \multicolumn{1}{c|}{-} & - \\
\multicolumn{1}{|c|}{CARPK \cite{hsieh2017drone}} & \multicolumn{1}{c|}{OC} & \multicolumn{1}{c|}{2017} & \multicolumn{1}{c|}{-} & \multicolumn{1}{c|}{1.4k} & \multicolumn{1}{c|}{1280$\times$720} & \multicolumn{1}{c|}{R} & - & \multicolumn{1}{c|}{-} & \multicolumn{1}{c|}{-} & - \\
\multicolumn{1}{|c|}{UAVDT \cite{UAVDT}} & \multicolumn{1}{c|}{OD, OT} & \multicolumn{1}{c|}{2018} & \multicolumn{1}{c|}{100} & \multicolumn{1}{c|}{80k} & \multicolumn{1}{c|}{1024$\times$540} & \multicolumn{1}{c|}{R} & - & \multicolumn{1}{c|}{-} & \multicolumn{1}{c|}{\checkmark} & \checkmark \\
\multicolumn{1}{|c|}{VisDrone \cite{visdrone}} & \multicolumn{1}{c|}{OD, OT} & \multicolumn{1}{c|}{2018} & \multicolumn{1}{c|}{263} & \multicolumn{1}{c|}{179.3k, static: 10k} & \multicolumn{1}{c|}{various} & \multicolumn{1}{c|}{R} & \checkmark & \multicolumn{1}{c|}{-} & \multicolumn{1}{c|}{-} & - \\
\multicolumn{1}{|c|}{DroneSURF \cite{kalra2019dronesurf}} & \multicolumn{1}{c|}{FRD} & \multicolumn{1}{c|}{2019} & \multicolumn{1}{c|}{200} & \multicolumn{1}{c|}{411.5k} & \multicolumn{1}{c|}{1280$\times$720} & \multicolumn{1}{c|}{R} & \checkmark & \multicolumn{1}{c|}{-} & \multicolumn{1}{c|}{-} & - \\
\multicolumn{1}{|c|}{AU-AIR \cite{bozcan2020air}} & \multicolumn{1}{c|}{OD} & \multicolumn{1}{c|}{2020} & \multicolumn{1}{c|}{8} & \multicolumn{1}{c|}{32.8k} & \multicolumn{1}{c|}{1920$\times$1080} & \multicolumn{1}{c|}{R} & \checkmark & \multicolumn{1}{c|}{-} & \multicolumn{1}{c|}{\checkmark} & - \\
\multicolumn{1}{|c|}{MOR-UAV \cite{MOR-UAV}} & \multicolumn{1}{c|}{MOR} & \multicolumn{1}{c|}{2020} & \multicolumn{1}{c|}{30} & \multicolumn{1}{c|}{10.9k} & \multicolumn{1}{c|}{various} & \multicolumn{1}{c|}{R} & - & \multicolumn{1}{c|}{-} & \multicolumn{1}{c|}{-} & - \\
\multicolumn{1}{|c|}{DOTA \cite{ding2021object}} & \multicolumn{1}{c|}{OD} & \multicolumn{1}{c|}{2021} & \multicolumn{1}{c|}{-} & \multicolumn{1}{c|}{11.3k} & \multicolumn{1}{c|}{various} & \multicolumn{1}{c|}{R} & - & \multicolumn{1}{c|}{-} & \multicolumn{1}{c|}{-} & - \\
\multicolumn{1}{|c|}{UAV-Human \cite{li2021uav}} & \multicolumn{1}{c|}{AR, PE, PR, ATR} & \multicolumn{1}{c|}{2021} & \multicolumn{1}{c|}{AR: 67.4k} & \multicolumn{1}{c|}{PE: 22.5k, PR: 41.3k, ATR: 22.3k} & \multicolumn{1}{c|}{1920$\times$1080} & \multicolumn{1}{c|}{R} & \checkmark & \multicolumn{1}{c|}{\checkmark} & \multicolumn{1}{c|}{-} & - \\
\multicolumn{1}{|c|}{SeaDroneSee \cite{varga2022seadronessee}} & \multicolumn{1}{c|}{OD, OT} & \multicolumn{1}{c|}{2022} & \multicolumn{1}{c|}{SOT: 208, MOT: 22} & \multicolumn{1}{c|}{OD: 5.6k, MOT: 54.1k} & \multicolumn{1}{c|}{various} & \multicolumn{1}{c|}{R} & \checkmark & \multicolumn{1}{c|}{-} & \multicolumn{1}{c|}{\checkmark} & - \\
\multicolumn{1}{|c|}{A212-Haze \cite{narayanan2022multi}} & \multicolumn{1}{c|}{IER, OD} & \multicolumn{1}{c|}{2022} & \multicolumn{1}{c|}{-} & \multicolumn{1}{c|}{1k} & \multicolumn{1}{c|}{1845$\times$1500} & \multicolumn{1}{c|}{R} & \checkmark & \multicolumn{1}{c|}{-} & \multicolumn{1}{c|}{\checkmark} & - \\ 
\multicolumn{1}{|c|}{DGTA \cite{kiefer2022leveraging}} & \multicolumn{1}{c|}{OD} & \multicolumn{1}{c|}{2022} & \multicolumn{1}{c|}{-} & \multicolumn{1}{c|}{VisDrone: 50k, SeaDroneSee: 100k, Cattle: 50k} & \multicolumn{1}{c|}{3840$\times$2160} & \multicolumn{1}{c|}{S} & \checkmark & \multicolumn{1}{c|}{-} & \multicolumn{1}{c|}{\checkmark} & - \\ \hline
\multicolumn{1}{|c|}{\textbf{Archangel-Real}} & \multicolumn{1}{c|}{OD} & \multicolumn{1}{c|}{2022} & \multicolumn{1}{c|}{69} & \multicolumn{1}{c|}{41.4k} & \multicolumn{1}{c|}{1304$\times$978} & \multicolumn{1}{c|}{R} & \checkmark & \multicolumn{1}{c|}{\checkmark} & \multicolumn{1}{c|}{\checkmark} & - \\
\multicolumn{1}{|c|}{\textbf{Archangel-Mannequin}} & \multicolumn{1}{c|}{OD} & \multicolumn{1}{c|}{2022} & \multicolumn{1}{c|}{598} & \multicolumn{1}{c|}{178.8k} & \multicolumn{1}{c|}{1920$\times$1080} & \multicolumn{1}{c|}{R} & \checkmark & \multicolumn{1}{c|}{\checkmark} & \multicolumn{1}{c|}{\checkmark} & - \\
\multicolumn{1}{|c|}{\textbf{Archangel-Synthetic}} & \multicolumn{1}{c|}{OD} & \multicolumn{1}{c|}{2022} & \multicolumn{1}{c|}{-} & \multicolumn{1}{c|}{4423.7k} & \multicolumn{1}{c|}{512$\times$512} & \multicolumn{1}{c|}{S} & \checkmark & \multicolumn{1}{c|}{\checkmark} & \multicolumn{1}{c|}{\checkmark} & \checkmark \\ \hline
\end{tabular}%
}

\bigskip
\resizebox{\textwidth}{!}{%
\begin{tabular}{|c|c||c|c||c|c||c|c|}
\hline
Notation & Description & Notation & Description & Notation & Description & Notation & Description \\ \hline
TF & Trajectory Forecasting & OT & Object Tracking & OD & Object Detection & AR & Action Recognition \\
OC & Object Counting & FRD & Face Recognition \& Detection & PE & Pose Estimation & MOR & Moving Object Recognition \\
PR & Person Re-identification & ATR & Attribute Recognition & SOT/MOT & Single/Multiple Object Tracking & IER & Image Enhancement \& Restoration \\ \hline
\end{tabular}%
}
\end{table*}

\subsection{UAV-based Object Detection Methods}

With the rapid development of generic object detection methods \cite{zaidi2022survey} and the aforementioned UAV-based object detection benchmarks (Tab.~\ref{tab:dataset-summary}), the detection accuracy of UAV-based object detectors has improved significantly over the past few years. In addition to common issues for generic object detection, UAV-based object detection has its own unique challenges \cite{wu2021deep}. In general, all of the challenges can be roughly divided into three categories. First, objects in UAV-based images are usually much smaller \cite{yu20201st}. Therefore, many solutions have been proposed to address this problem to date. For example, Liu et al. \cite{liu2021hrdnet} proposed HRDNet that fused information from both high- and low-resolution inputs to simultaneously preserve features of small objects and maintain computational costs. Similarly, Liu et al. \cite{liu2020small} introduced a multi-branch and parallel structure (MPFPN) to extract more powerful features for tiny object detection. Besides the scale of an object, target objects in UAV-based datasets are usually crowded and sparsely distributed, reducing both the accuracy and efficiency of an object detector. Thus, Yang et al. \cite{yang2019clustered} proposed ClusDet that performed object cluster proposal first before detecting objects. Finally, UAV-based datasets contain many UAV-specific nuisances \cite{NDFT}, such as varying UAV altitudes and viewing angles. These nuisances cause tremendous variations in object appearances, causing degraded detection performance. To address this issue, Wu et al. \cite{NDFT} proposed to adopt adversarial training to learn domain-robust features from UAV-specific nuisances coarsely annotated by the authors. In this paper, we posit that UAV-based object detection can be further enhanced by providing UAV-based benchmarks with a set of fine-grained metadata, such as that contained in \textit{Archangel}.

In addition to improving the detection accuracy, reducing the computational cost to achieve real-time on-board processing is also very important for UAV-based object detection approaches. One way to improve latency is to skip unnecessary computation. For instance, Ammour et al. \cite{small-region} proposed to extract candidate regions of target objects first via over-segmentation. After that, only windows around the candidate regions were sent to the pre-trained CNN and linear SVM for feature extraction and classification. Another way of reducing computational overhead is to use more efficient one-stage object detectors, such as YOLO \cite{bochkovskiy2020yolov4, yolo}, RetinaNet \cite{8417976}, CenterNet \cite{9010985}, and EfficientDet \cite{9156454}. These one-stage object detectors directly classify and locate objects without generating region proposals, resulting in improved latency. As an example, Liu et al. \cite{uav-yolo} adapted the original network architecture of YOLO by making it more suitable for UAV-based object detection. In this paper, we also utilize YOLOv5 \cite{yolo} for all the experiments due to the advantage of its low complexity (Tab.~\ref{tab:yolo}).

\begin{table}
\centering
\caption{Complexity of the YOLOv5 models \cite{yolo} used in this study.}
\label{tab:yolo}
\begin{tabular}{|c||c|c|}
\hline
Model             & \#Parameters (M) & FLOPs (G) \\ \hline
\textit{YOLOv5n6} & 3.1              & 4.3       \\ 
\textit{YOLOv5s6} & 12.3             & 16.2      \\ 
\textit{YOLOv5m6} & 35.3             & 49.1      \\ \hline
\end{tabular}
\end{table}

\section{The Archangel Dataset}
\begin{figure}[h]
\centering
\includegraphics[width=\columnwidth]{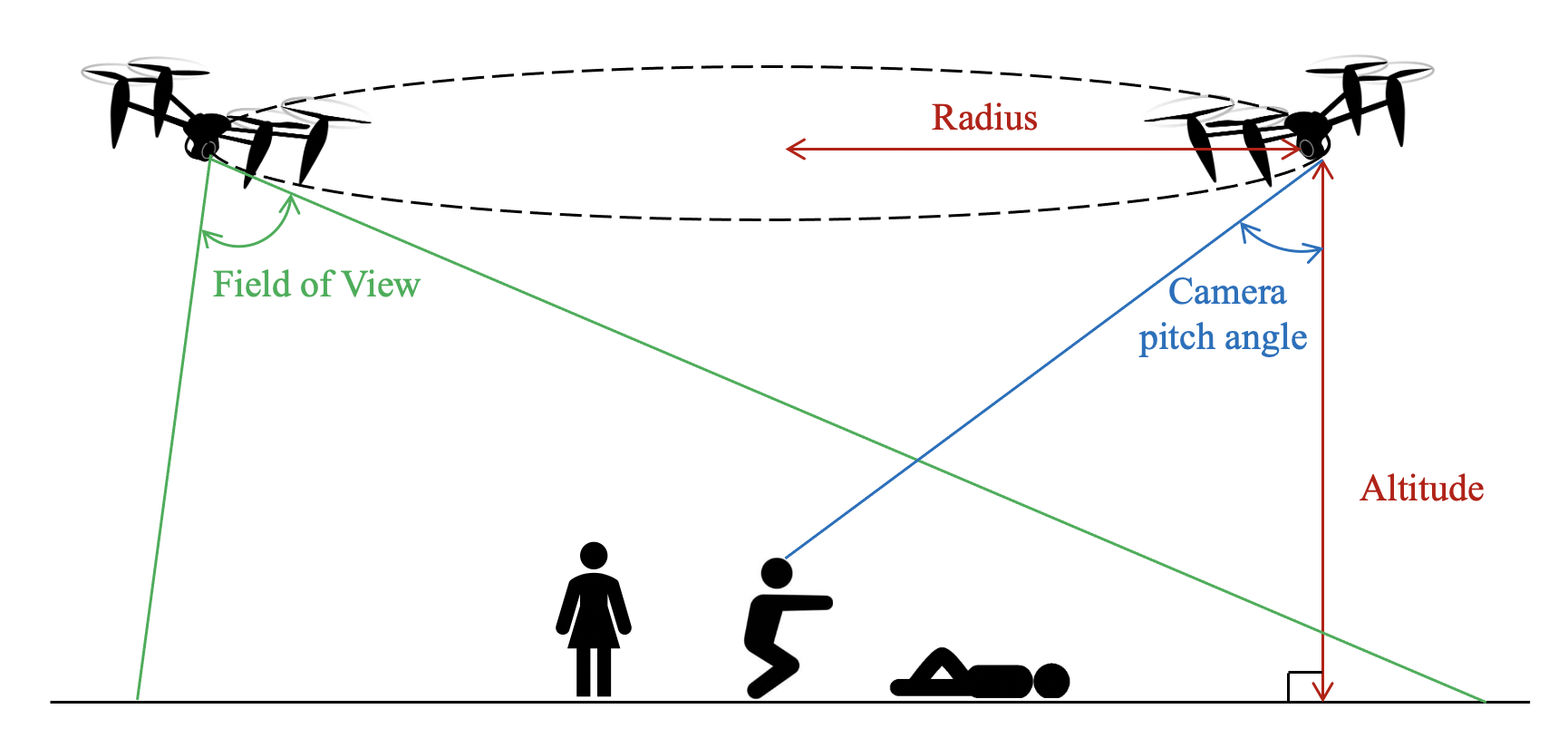}
\caption{Illustration of the data collection process for \textit{Archangel}. For each data collection, a number of objects (real people, mannequins or virtual characters) on the ground were captured by a camera mounted on a UAV (real or simulated). Each of the objects was in one of the three defined poses (\textit{stand}, \textit{kneel}, and \textit{prone}). The UAV circled around the objects at a predefined altitudes and radii of rotation circles.}
\label{fig:collection}
\end{figure}

The data collection process for \textit{Archangel} is illustrated in Fig.~\ref{fig:collection} and a brief comparison of the three sub-datasets is provided in Tab.~\ref{tab:dataprop}. In the following, we will go through the data collection process of each sub-dataset in detail. 

\begin{table*}
\centering
\caption{Comparison of the three sub-datasets comprising Archangel.}
\label{tab:dataprop}
\resizebox{\textwidth}{!}{
\begin{tabular}{|c||c|c|c|c|c|c|} 
\hline
\textbf{Archangel} & Image Size & Targets & Altitude (m) & Radius (m) & Field of View & Camera Pitch Angle \\ [0.5ex] 
\hline
\textbf{Synthetic} & 512$\times$512 & Virtual characters & [5-80] increment by 5 & [5-80] increment by 5 & 22.5$\degree$ & various  \\
\textbf{Mannequin} & 1920$\times$1080 & Mannequins, vehicles & [15-50] increment by 5 & [15-50] increment by 5 &  120$\degree$ &  45$\degree$ \\
\textbf{Real} & 1304$\times$978 & Real people & [15-50] increment by 5 & [20-50] increment by 5 &  45$\degree$  & 22.5$\degree$, 45$\degree$, 67.5$\degree$ \\
\hline
\end{tabular}
}
\end{table*}

\subsection{Archangel-Mannequin}

\noindent{\bf Target Objects.} During this data collection, a group of mannequins were used as human surrogates primarily due to the safety guidelines of the test facility and the difficulty of asking humans to maintain certain strenuous poses, such as \textit{prone} and \textit{squat}, for a long period of time. The mannequins were dressed in casual outfits and positioned in three different poses (i.e., \textit{stand}, \textit{kneel}, and \textit{prone}). Hence, the distribution gap between human attributes and those of mannequins' is not noticeably large. Additionally, the dataset also includes a small group of various types of civilian vehicles as targets, such as sports utility vehicles (SUVs), minivans, and sedans. Each target in the dataset was labeled as \textit{mannequin-standing}, \textit{-kneeling}, \textit{-prone} or \textit{civilian vehicles}.\smallskip

\noindent{\bf Data Collection.} The imagery was captured using a contractor-built UAV equipped with an onboard electro-optical (EO) camera (ELP-USBFHD01M-L21) with a 1920$\times$1080 pixel array and a lens with approximately 120\degree \:field-of-view (FOV). The UAV camera was pitched forward by 45\degree\ relative to level flight. During the course of multiple UAV flights, the UAV operated over a wide range of altitudes and radii of rotation circles while keeping the camera pointed inward toward the targets and circling a central point. Both the altitude and radius of the rotation circle were varied from 15-50 meters in 5-meter increments. Since the target objects were stationary and the camera pitch angle was constant, the target objects were spread across different regions of the camera's FOV, resulting in different view angles.

\subsection{Archangel-Synthetic}

\noindent{\bf Motivation.} While \textit{Archangel-Mannequin} provides valuable aerial imagery with pose and position metadata well suited for UAV-based human detection, the imaging conditions of this data collection were limited. Furthermore, \textit{Archangel-Mannequin} did not incorporate some important factors, such as various human appearances/attributes, extended ranges of UAV altitudes and radii of rotation circles, and different illumination conditions. To overcome these restrictions, a large-scale synthetic imagery (i.e., \textit{Archangel-Synthetic}) dataset containing multiple virtual humans in the same poses as that used in \textit{Archangel-Mannequin} was generated using the Unity game engine \cite{unity} to augment \textit{Archangel}.\smallskip 

\noindent{\bf Data Generation and Labeling.} In the Unity-based simulation, a 3D scene is constructed using a terrain asset (i.e., background) and one or more target assets (i.e., virtual characters in different outfits and poses). For the current version of \textit{Archangel-Synthetic}, we use only a simple terrain model (i.e., desert), but we plan to integrate more complex terrain models in future work. 

For each target asset, we first created a Unity project and configured the lighting and camera parameters in a virtual 3D environment. A Unity terrain asset (i.e., the desert background) and the target asset were then added to the 3D environment. After the 3D environment was configured as above, a C\# script was then used to control the position and viewing angle of the camera as it circled around the target. At each step, the camera was pointed at the center of the target. The script was iterated to encompass the whole range of UAV camera altitudes, the radii of the circles, and the camera viewing angles relative to the target, thus producing imagery captured at various camera-to-target distances and camera pitch angles. Additionally, the sun angle was varied to generate synthetic images captured at different times of a day with corresponding illumination conditions. This resulted in large-scale synthetic imagery with significant target pose, scale, and illumination variations.

To synthesize images and annotations from the virtual 3D environment constructed above, we used an open source software asset, Image Synthesis for Machine Learning \cite{U3DC}. Specifically, the software produced an image segmentation mask where each target object in a synthetic image was assigned a unique scalar value. To generate the bounding box annotations for each target, a Python script was used to parse each segmentation mask, identify each target object, and measure the center, width, and height of the tightest bounding box encompassing the target. Additionally, the target category, the camera position, the target orientation relative to the camera, the camera-to-target distance, the camera pitch angle, and the number of pixels inside the segmentation mask were recorded in a single JavaScript Object Notation (JSON) file for each trial.\smallskip

\noindent{\bf Properties.} \textit{Archangel-Synthetic} includes mountainous desert terrain and eight different virtual characters, each in three different poses (i.e., \textit{stand}, \textit{squat}, and \textit{prone}). In a single trial of synthetic data generation (i.e., a virtual character with a certain pose), both the altitude of the camera and the radius of the rotation circle were varied from 5-80 meters in 5-meter increments. Additionally, the camera viewing angle relative to the character was varied from 0\degree-358\degree\ in 2\degree\ increments and four different sun angles were simulated. This resulted in over 4.4M images included in \textit{Archangel-Synthetic}. Each image contains 512$\times$512 pixels with horizontal and vertical fields-of-view of 22.5\degree.

\subsection{Archangel-Real}

\noindent{\bf System Design.} The dataset was collected with an ARL-designed UAV platform called the Dawn Dove (D2). The D2 is a re-configurable UAV, with the ability to shift the center of gravity by adjusting arm angles, arm placement, battery, sensor payload, and on-board processor location. It is composed of a combination of 3D printed polyethylene terephthalate glycol (PETG) and carbon fiber infused nylon, and traditional carbon fiber parts. It can carry various types of sensor payloads and on-board processors. It has an approximate payload capacity of 1.5 lbs and an approximate flight time of 8 minutes. For this data collection, the sensor payload consisted of a UI-3250ML-C-HQ EO camera with an Edmund Optics 6mm/F1.4 lens and a FLIR Boson 640 8.7 mm IR camera. The cameras were co-located on the front of the D2 and the EO camera’s image was cropped to match the FOV of the FLIR Boson (50\degree\ HFOV). The on-board processor was an NVIDIA Xavier NX with a 1 TB NVMe SSD for additional data storage.\smallskip

\noindent{\bf Data Collection.} In this dataset, the targets consisted of real people wearing civilian clothing in three different poses: \textit{stand}, \textit{kneel}, and \textit{prone}. The data collection process involved having the D2 fly circles at radii ranging from 20-50 meters, at intervals of 5 meters, and at altitudes ranging from 15-50 meters, at intervals of 5 meters. The camera angle relative to level flight was manually adjusted between flights from -22.5\degree, -45\degree, and -67.5\degree to ensure the targets remained within the FOV of the cameras. In total, 52 circles were flown around the targets. To fly the circles, custom Robot Operating System (ROS) based autonomy code was used, along with a custom Python-based graphical user interface (GUI), which communicated with the UAV. From the ground control station (GCS), the target GPS location, circle radius, altitude, maximum velocity, and file name were entered into the GUI. Once the circle parameters were entered, the D2 was manually armed, launched, and switched over to “offboard mode” which passed control of the UAV to the GCS. The GCS then commanded the UAV to perform the autonomous circle. Once complete, the next circle’s parameters were entered into the GUI and sent to the UAV while still in the air. This was repeated each flight until the UAV had to be brought back down to replace the battery. In addition to stationary targets, a few circles were also flown where the people walked, jogged, crawled, and waved. Note that \textit{Archangel-Real} involves human subjects as UAV-based detection instances. However, an Institutional Review Board (IRB) approval was exempted since one cannot identify individuals in the dataset.

\subsection{Importance of the Camera Parameters}

Before moving on to the data analysis section, we want to highlight the importance of revealing the camera parameters used in the data collection. Note that the scale of human instances in UAV-based object detection datasets, such as \textit{Archangel}, is strongly influenced by the camera parameters used in the data collection, including FOV, pixel-array size, and pitch angles, in addition to the UAV altitude and radius of rotation circle. Hence, the detection results can vary greatly when using different camera parameters. However, all the conclusions derived from the following data analysis of \textit{Archangel} can still be applied to other UAV-based datasets using different camera parameters through extrapolation. That is, the performance gap can be easily calibrated by adjusting the scale of human instances if the camera parameters and the original object size are known \textit{a priori}.

\section{Experimental Setup}
\label{sec:setup}
\begin{figure}
\centering
\includegraphics[width=\columnwidth]{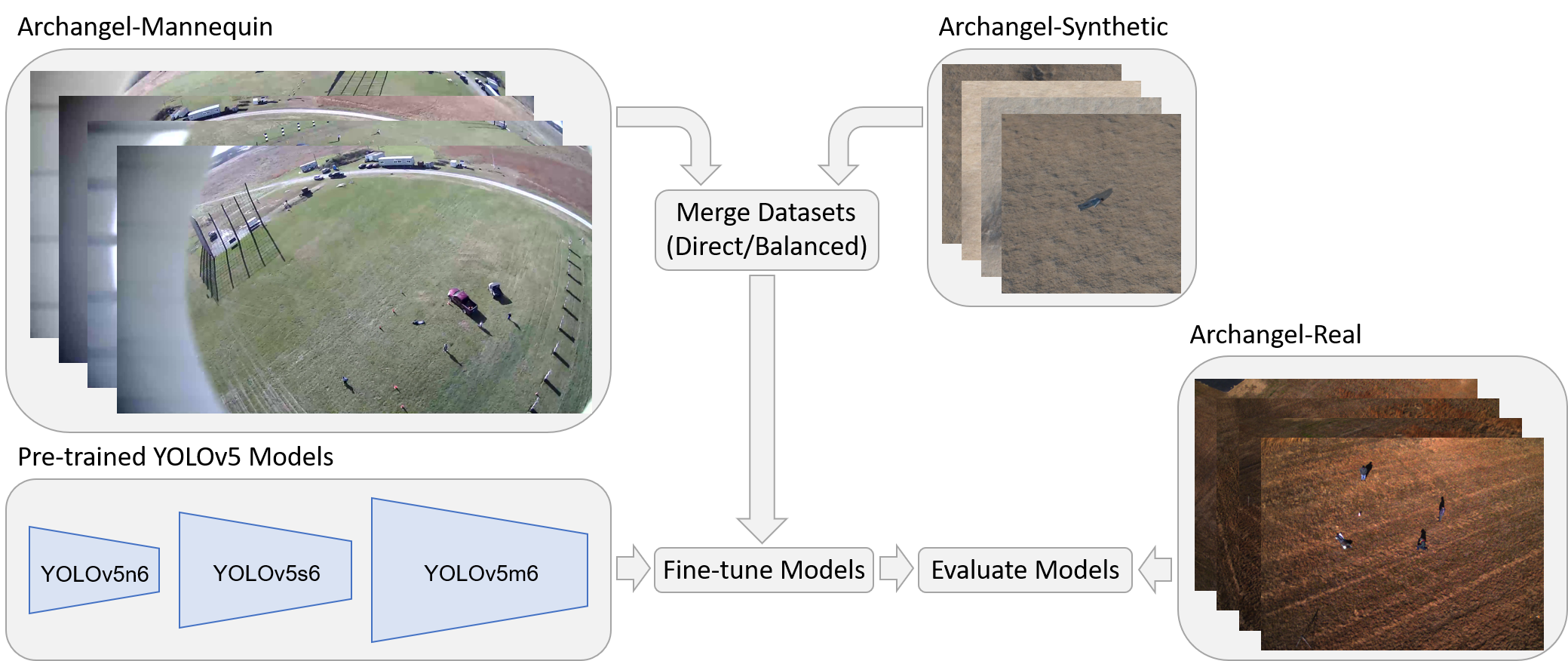}
\caption{Overview of the designed experimental flow. The pre-trained YOLOv5 models with various complexity were fine-tuned on different hybrid sets of UAV-based real and synthetic data and evaluated on a hold-out UAV-based dataset.}
\label{fig:overview}
\end{figure}

\noindent{\bf Overview.} In this paper, we designed a series of experiments based on the flow shown in Fig.~\ref{fig:overview}. In brief, for each experiment, we selected one of the three pre-trained YOLOv5 models (Tab.~\ref{tab:yolo}) and fine-tuned the model on varying amounts of UAV-based real (i.e., \textit{Archangel-Mannequin}) and synthetic (i.e, \textit{Archangel-Synthetic}) data. We then evaluated the model on a sequestered UAV-based dataset (i.e., \textit{Archangel-Real}). Based on the results, the designed experimental flow can provide valuable insights into optimizing UAV-based object detectors with hybrid sets of real and synthetic data.\smallskip

\noindent{\bf Datasets.} Note that acquiring the best performance for an UAV-based object detector is not the primary purpose of this study. Thus, we subsampled each of the three sub-datasets of \textit{Archangel} to explore optimal strategies for fine-tuning or evaluating models:
\begin{enumerate}
    \item \textit{Archangel-Mannequin}: The dataset consists of video clips collected in 11 UAV flight trials. In this paper, we carefully split the dataset into two subsets so that each covered the whole range of the UAV positions covered during the entire data collection. The video clips collected in 6 of the 11 trials (i.e., Trial-5, 6, 8, 9, 10, 11) were used for evaluating models. The rest (i.e., Trial-1, 2, 3, 4, 7) were used for fine-tuning models. All the video clips were uniformly subsampled at 3 fps. This resulted in two sets of frames, \textit{Arch-Mann-FT37}, containing 6.7k frames for fine-tuning models, and \textit{Arch-Mann-Eval}, containing 11.2k frames for evaluating models. \textit{Arch-Mann-FT37} is named based on the amount of data it has compared to the entire \textit{Archangel-Mannequin} in terms of percentage (i.e., 37\%).
    \item \textit{Archangel-Real}: Similarly, we uniformly subsampled the video clips in \textit{Archangel-Real} at 1 fps. This resulted in a set of frames, named as \textit{Arch-Real-Eval}, containing 4.1k frames for evaluating models. 
    \item \textit{Archangel-Synthetic}: Only one virtual character in all of the three poses was used. For each UAV position, only one of the four sun angles was randomly selected. Additionally, instead of using all the UAV positions, we uniformly sampled images across each rotation circle in 60\degree \:increments. This resulted in a set of images, named as \textit{Arch-Syn-FT}, containing 4.6k images for fine-tuning models.
\end{enumerate}

Each of the three sub-datasets has its own unique usage in this study. In general, \textit{Archangel-Real} serves as the primary UAV-based benchmark for measuring detection accuracy. \textit{Archangel-Mannequin} can be viewed as the real UAV-based fine-tuning dataset for adapting the detection models to the target UAV-based domain. Although it includes mannequins instead of real humans, fine-tuning on this dataset is shown to be effective in the following sections. \textit{Archangel-Synthetic}, on the other hand, is used as the synthetic version of the UAV-based fine-tuning dataset, which can be combined with \textit{Archangel-Mannequin} to further optimize the models.\smallskip

\noindent{\bf Evaluation.} We utilized standard AP50, the average precision with an IOU (Intersection of Union) threshold of 0.5, as the metric to measure the performance of each object detector. Moreover, AP50 was computed for each pose respectively. As more than one pose may exist in a single image, to obtain the performance for only one certain pose, the other two poses were ignored during the evaluation process.\smallskip

\noindent{\bf Implementation Details.} The official repository of YOLOv5 \cite{yolo} was used for both fine-tuning and evaluating models. If not specified otherwise, the default hyperparameters were adopted. The input images were rescaled (i.e., \textit{imgsz}=1280) first before being fed into all the models. During fine-tuning, the backbone for each model was frozen (i.e., \textit{freeze}=10) to prevent the model from easily over-fitting. We fine-tuned each model for 20 epochs with a batch size of 16 on a server with 4 NVIDIA GeForce RTX 2080 TI GPUs. During the evaluation, we set the confidence threshold to be 0.05.

\section{Results}
\label{sec:results}
\begin{figure}
\centering
\includegraphics[width=\columnwidth]{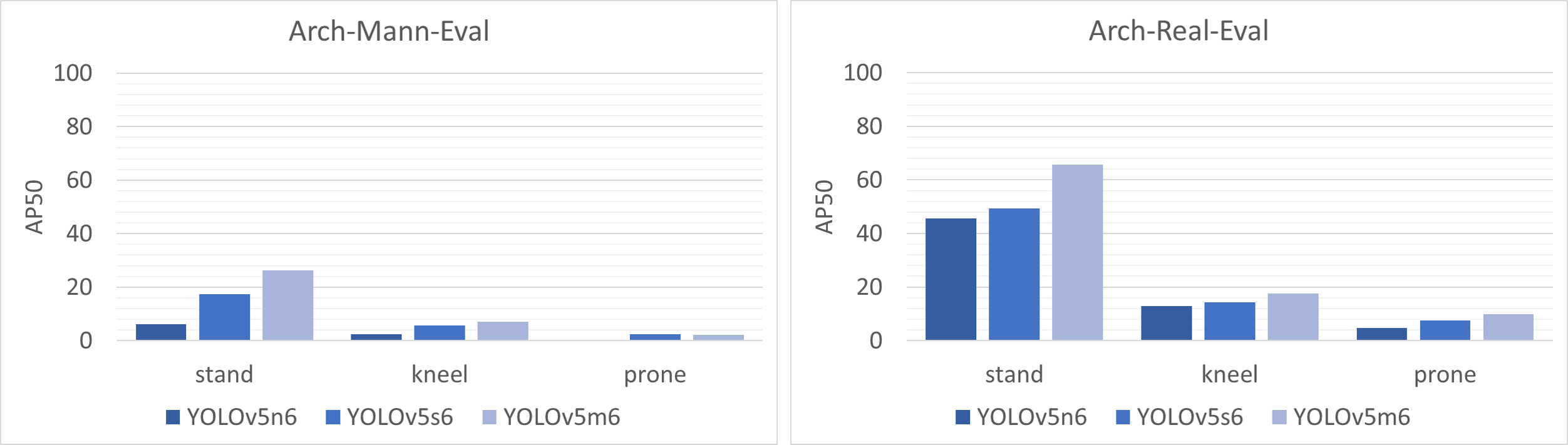}
\caption{AP50 of the pre-trained YOLOv5 models evaluated on \textit{Arch-Mann-Eval} and \textit{Arch-Real-Eval}.}
\label{fig:pretrained}
\end{figure}

\noindent{\bf The Performance of Pre-trained Models.} To begin with, we evaluated the three pre-trained YOLOv5 models on \textit{Arch-Mann-Eval} and \textit{Arch-Real-Eval}. The models were pre-trained on MS-COCO, a representative ground-based dataset. The results are shown in Fig.~\ref{fig:pretrained}. From the results, we can gain several useful insights on UAV-based object detection. First, larger pre-trained models achieved better accuracy across all the evaluation datasets and poses. One possible reason for this is that larger models, compared with smaller ones, can explore better and find more powerful features for classification and detection \cite{brutzkus2019larger}. Although it implies that we can get higher accuracy by using larger pre-trained models, such larger models may not fit well on small UAV platforms with computational constraints.

Another trend we can observe is that the pre-trained models had much better accuracy on \textit{stand}. It is mainly because that the dataset used for pre-training models (i.e., MS-COCO) contains significantly more human instances in \textit{stand} (i.e., 84.53\%) \cite{coco-pose}. In other words, it is impractical to directly use detectors pre-trained on standard datasets, especially in unusual scenarios where we need to detect people in uncommon positions, such as search and rescue in disaster relief. It is worth mentioning that the pre-trained YOLOv5 models performed much worse on \textit{Arch-Mann-Eval} than on \textit{Arch-Real-Eval}. That is mainly because \textit{Arch-Mann-Eval} contains some other objects, such as traffic cones and fiducials, which are easily misclassified as humans when captured by the pre-trained detectors at high altitudes \cite{realarchangel}.\smallskip

\noindent{\bf Fine-tuning Models on a Real UAV-based Dataset: Arch-Mann-FT37.} As we have discussed, most ground-based datasets used to fine-tune models usually lack UAV-specific samples for the models to learn from, such as human instances in non-standing positions captured from various camera viewing angles and altitudes. Therefore, we allowed the pre-trained YOLOv5 models to acquire such knowledge by fine-tuning the models on \textit{Arch-Mann-FT37}. Moreover, given the significant challenges of collecting and annotating UAV-based datasets \cite{visdrone} and the lack of existing large-scale UAV-based object detection benchmarks, we explored the idea of fine-tuning models in the small-data regime \cite{bornschein2020small}. More precisely, we subsampled the original fine-tuning dataset and created several smaller subsets for fine-tuning, which contained much less data (i.e., \textit{Arch-Mann-FT20}, \textit{Arch-Mann-FT10}, \textit{Arch-Mann-FT5} and \textit{Arch-Mann-FT2}). We followed the same naming strategy as \textit{Arch-Mann-FT37} for the extra fine-tuning datasets.

\begin{figure*}
\centering
\includegraphics[width=\textwidth]{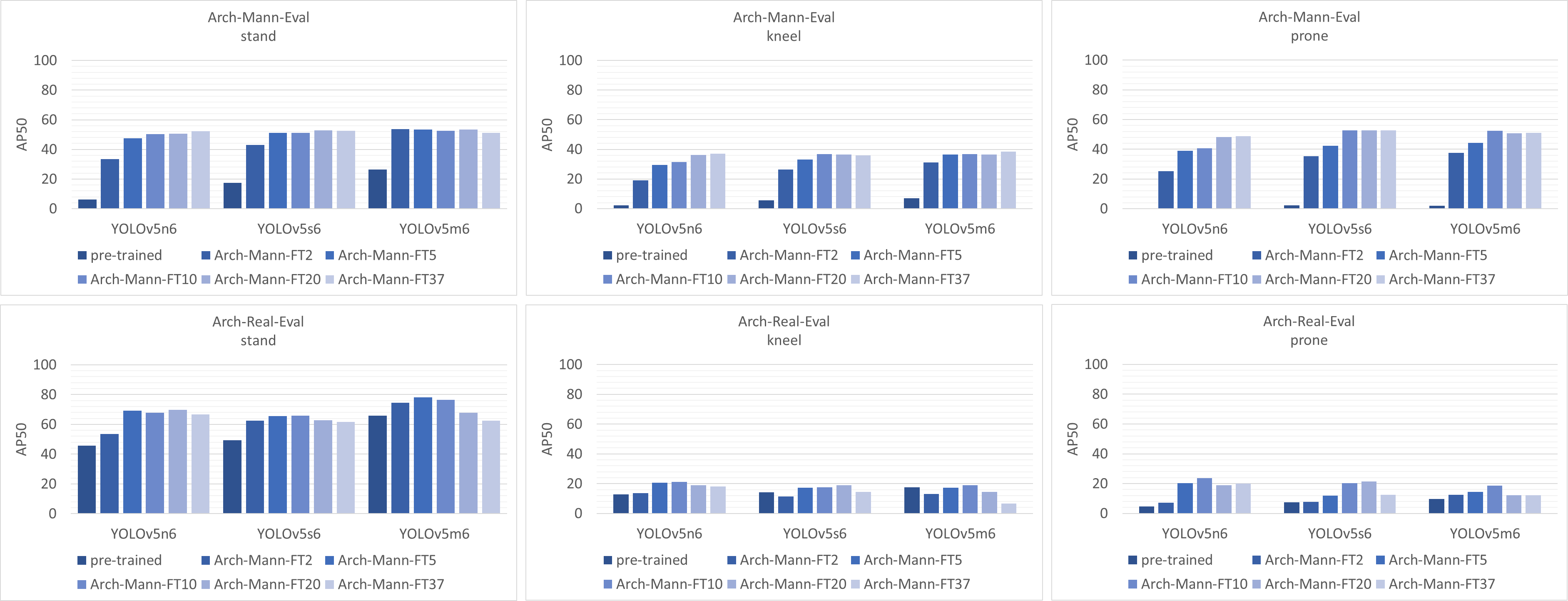}
\caption{AP50 of the YOLOv5 models fine-tuned on the different subsets of \textit{Arch-Mann-FT37}. Each model was evaluated on \textit{Arch-Mann-Eval} (top) and \textit{Arch-Real-Eval} (bottom).}
\label{fig:ft_mann}
\end{figure*}

The results are presented in Fig.~\ref{fig:ft_mann}. We would like to highlight the importance of having an evaluation dataset with different characteristics from the fine-tuning dataset. As we fine-tuned the models on data from \textit{Arch-Mann-FT37}, we could improve their detection accuracy on a similar evaluation dataset such as \textit{Arch-Mann-Eval}. However, the models fine-tuned on too much data from \textit{Arch-Mann-FT37} tended to perform worse on \textit{Arch-Real-Eval}. We argue that this was because the models started to learn certain dataset-specific features from the fine-tuning dataset, adversely affecting the models' generalization capability to unseen datasets, such as the evaluation dataset in our case. In practice, ML models embedded into UAVs are usually deployed to new environments unseen during training. Thus, in the following experiments, we chose to fine-tune models on data from \textit{Arch-Mann-FT37} and \textit{Arch-Syn-FT} but evaluate them on \textit{Arch-Real-Eval}.\smallskip

\noindent{\bf Fine-tuning Models on Both Real and Synthetic Datasets: Arch-Mann-FT37 and Arch-Syn-FT.} One of the significant advantages of \textit{Archangel} is that it contains both real and synthetic subsets acquired from similar imaging conditions in data collections and synthetic rendering, respectively. Hence, investigating the effect of augmenting the original UAV-based fine-tuning datasets with UAV-based synthetic data is another major topic for this study. To do so, we fine-tuned the pre-trained YOLOv5 models on \textit{Arch-Syn-FT}. Additionally, a well-known concern about learning from synthetic data is that, compared with real data, synthetic data usually contains much less variations in appearances/attributes of objects or structures of scenes \cite{seib2020mixing}. Therefore, instead of fine-tuning models only on \textit{Arch-Syn-FT}, we also explored the idea of multi-source learning \cite{zhou2022domain}, constructing multiple hybrid fine-tuning datasets by directly merging \textit{Arch-Syn-FT} with all the subsets of \textit{Arch-Mann-FT37} used in the previous experiment. 

\begin{figure*}
\centering
\includegraphics[width=\textwidth]{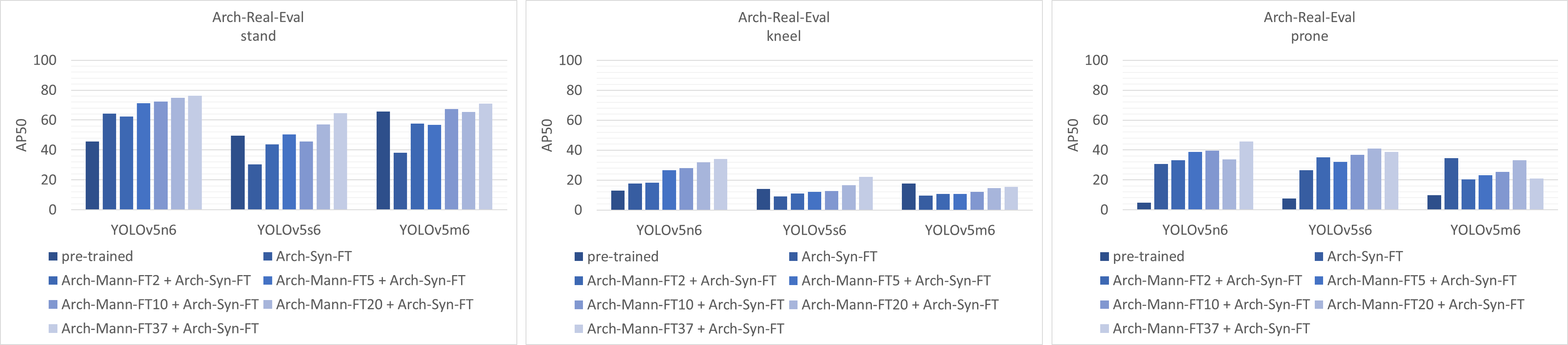}
\caption{AP50 of the YOLOv5 models fine-tuned on the various hybrid sets constructed by \textit{Arch-Syn-FT} and the different subsets of \textit{Arch-Mann-FT37}.}
\label{fig:ft_mann_syn}
\end{figure*}

The results are shown in Fig. \ref{fig:ft_mann_syn}. For \textit{prone}, a rarely seen pose in the pre-training dataset, the detection accuracy of the pre-trained models was very low. After fine-tuning the models on the various hybrid subsets of \textit{Arch-Syn-FT} and \textit{Arch-Mann-FT37}, the detection accuracy continually increased with few exceptions. For \textit{stand}, the pre-trained models performed much better than they did for \textit{prone} as expected, but fine-tuning on the hybrid sets of the real and synthetic data still provided much improvement over the pre-trained models, as clearly observed from the detection accuracy of \textit{YOLOv5n6}. Similar observations can be made for \textit{kneel}.

\begin{figure*}
\centering
\includegraphics[width=\textwidth]{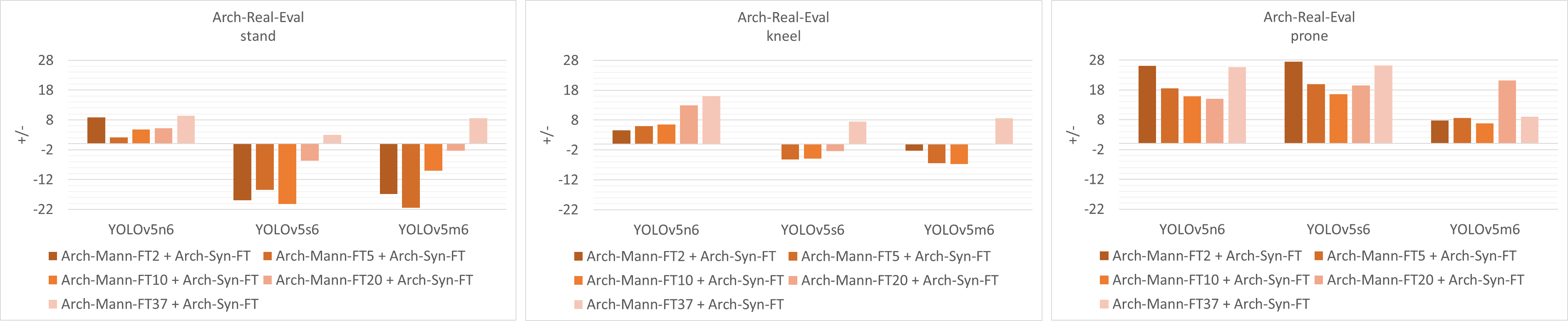}
\caption{AP50 improvement of the YOLOv5 models after adding \textit{Arch-Syn-FT} to the original fine-tuning datasets based on the different subsets of \textit{Arch-Mann-FT37}.}
\label{fig:ft_mann_syn_diff}
\end{figure*}

We now compare the results shown in Fig.~\ref{fig:ft_mann} with the ones shown in Fig.~\ref{fig:ft_mann_syn} to highlight the effect of adding the synthetic data to the fine-tuning dataset based only on the subsets of \textit{Arch-Mann-FT37}. The results are shown in Fig.~\ref{fig:ft_mann_syn_diff}. For \textit{prone}, we can observe significant performance improvement across all the models and hybrid subsets for fine-tuning after introducing the synthetic data into the fine-tuning datasets, compared with the results of fine-tuning on data from \textit{Arch-Mann-FT37} only. One explanation for the above finding is that most of the human instances in the pre-training dataset are in certain upright positions, including \textit{stand} and \textit{kneel}. Therefore, adding synthetic characters in \textit{prone} captured from various camera viewing angles and UAV altitudes to the fine-tuning dataset can effectively aid the models to learn how to detect humans in \textit{prone}. More in-depth analysis of this issue will be discussed in the ablation study (Sec.~\ref{sec:ablation}). 

Additionally, \textit{Arch-Syn-FT}, which includes only one virtual character and one type of background, might be too simple for larger models, such as \textit{YOLOv5s6} and \textit{YOLOv5m6} in our case, to learn from, causing them to overfit to the fine-tuning dataset. A proof of this assumption was that using \textit{Arch-Syn-FT} along with the subsets of \textit{Arch-Mann-FT37} to fine-tune \textit{YOLOv5s6} and \textit{YOLOv5m6} significantly decreased their performance on \textit{stand} and \textit{kneel}, especially when we included fewer data from \textit{Archangel-Mann-FT37} (Fig.~\ref{fig:ft_mann_syn_diff}). On the other hand, fine-tuning \textit{YOLOv5n6} on \textit{Arch-Syn-FT} along with the subsets of \textit{Arch-Mann-FT37} did not have such a negative effect. As a result, in the following ablation study, we focused on analyzing the results of \textit{YOLOv5n6} once we included \textit{Arch-Syn-FT} in the fine-tuning dataset.

\section{Ablation Study}
\label{sec:ablation}
\noindent{\bf Adjusting the Size of the Synthetic Dataset: Arch-Syn-FT.} We have demonstrated the effects of directly combining the various subsets of \textit{Arch-Mann-FT37} with the same set of synthetic data (i.e., \textit{Arch-Syn-FT}) for fine-tuning models (Fig.~\ref{fig:ft_mann_syn}~and~\ref{fig:ft_mann_syn_diff}). In this section, we are interested in further exploring the outcome of using different amounts of synthetic data to fine-tune models. Notably, we aim to investigate whether a "balanced" fine-tuning dataset, which contains the same amount of real and synthetic data, is better than its "unbalanced" counterpart, in which the ratio of the synthetic data to the real data varies due to the use of a fixed set of the synthetic data across all the hybrid fine-tuning datasets.

To achieve this goal, for each real fine-tuning dataset (i.e., \textit{Arch-Mann-FT37}, \textit{Arch-Mann-FT20}, \textit{Arch-Mann-FT10}, \textit{Arch-Mann-FT5}, and \textit{Arch-Mann-FT2}), the corresponding amount of data was randomly sampled from \textit{Arch-Syn-FT} to match the real fine-tuning dataset. Namely, if the real fine-tuning dataset was smaller than \textit{Arch-Syn-FT}, a subset of \textit{Arch-Syn-FT} was randomly selected and combined with the real fine-tuning dataset to form a balanced fine-tuning dataset. Similarly, if the real fine-tuning dataset was larger, a random subset of \textit{Arch-Syn-FT} was duplicated before the combination. We denote the synthetic fine-tuning dataset as \textit{Arch-Syn-FT-B} if the aforementioned data balancing procedure has been conducted.

\begin{figure*}
\centering
\includegraphics[width=\textwidth]{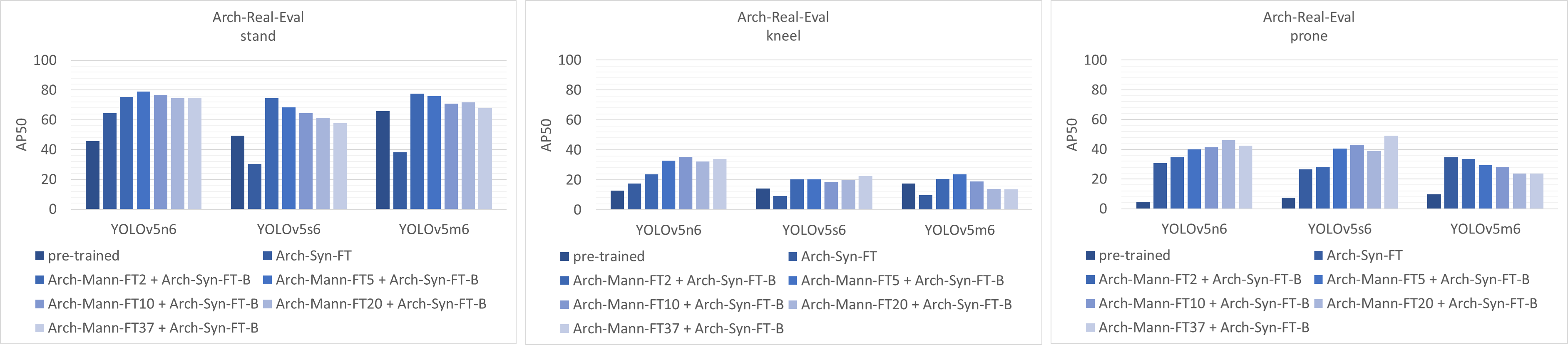}
\caption{AP50 of the YOLOv5 models fine-tuned on the various balanced hybrid sets constructed by \textit{Arch-Syn-FT-B} and the different subsets of \textit{Arch-Mann-FT37}.}
\label{fig:ft_mann_syn_b}
\end{figure*}

\begin{figure*}
\centering
\includegraphics[width=\textwidth]{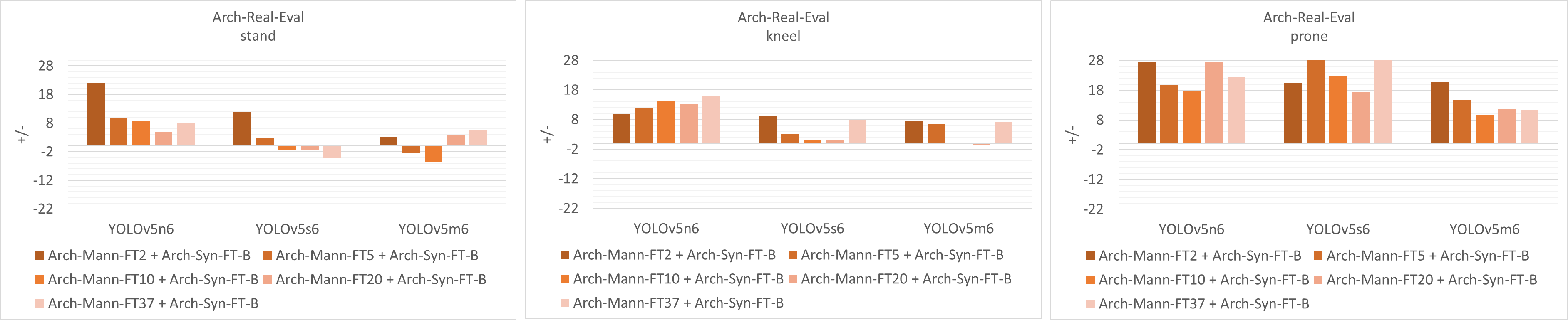}
\caption{AP50 improvement of the YOLOv5 models after adding \textit{Arch-Syn-FT-B} to the original fine-tuning datasets based on the different subsets of \textit{Arch-Mann-FT37}.}
\label{fig:ft_mann_syn_b_diff}
\end{figure*}

\begin{figure*}
\centering
\includegraphics[width=\textwidth]{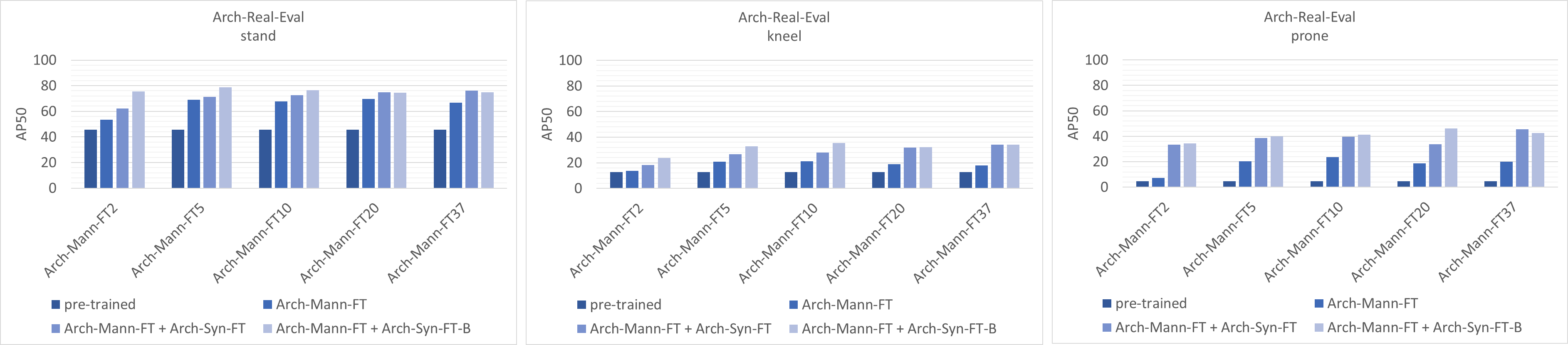}
\caption{AP50 comparison of \textit{YOLOv5n6} fine-tuned on the different hybrid sets of \textit{Arch-Syn-FT}, \textit{Arch-Syn-FT-B}, and the different subsets of \textit{Arch-Mann-FT37}.}
\label{fig:ft_mann_syn_b_nano}
\end{figure*}

The results are shown in Fig.~\ref{fig:ft_mann_syn_b},~\ref{fig:ft_mann_syn_b_diff}~and~\ref{fig:ft_mann_syn_b_nano}. Comparing Fig.~\ref{fig:ft_mann_syn_diff} with Fig.~\ref{fig:ft_mann_syn_b_diff}, we can observe that the negative effect of fine-tuning larger models (i.e., \textit{YOLOv5s6} and \textit{YOLOv5m6}) on hybrid fine-tuning sets largely decreases or even diminishes. Moreover, the improvement becomes more significant, especially for \textit{YOLOv5n6}. From Fig.~\ref{fig:ft_mann_syn_b_nano}, it is shown that fine-tuning \textit{YOLOv5n6} on the balanced hybrid sets of the real and synthetic data can always be at least on par with the best performance setting for fine-tuning. These findings indicate that the proportion of each data to the whole fine-tuning dataset significantly impacts the model's fine-tuning performance.\smallskip

\noindent{\bf Leaving One Pose Out from the Synthetic Dataset: Arch-Syn-FT.} We have claimed that fine-tuning models with synthetic human instances are particularly important for \textit{prone}, a pose rarely seen in the original training data. In this section, we would like to provide more evidence to support this statement. Specifically, we used \textit{YOLOv5n6} for this set of experiments since it showed less sign of overfitting when fine-tuning on \textit{Arch-Syn-FT} (Fig.~\ref{fig:ft_mann_syn}). Additionally, the technique of dataset balancing was adopted due to its positive effect on fine-tuning models (Fig.~\ref{fig:ft_mann_syn_b}~and~\ref{fig:ft_mann_syn_b_diff}). We followed a similar procedure to generate each hybrid fine-tuning dataset, except that each time one of the poses, \textit{stand}, \textit{kneel}, or \textit{prone}, was excluded in advance from \textit{Arch-Syn-FT}, resulting in a "leave-one-pose-out" fine-tuning dataset, \textit{Arch-Syn-FT-NoSt}, \textit{Arch-Syn-FT-NoKn}, or \textit{Arch-Syn-FT-NoPr}, respectively.

\begin{figure*}
\centering
\includegraphics[width=\textwidth]{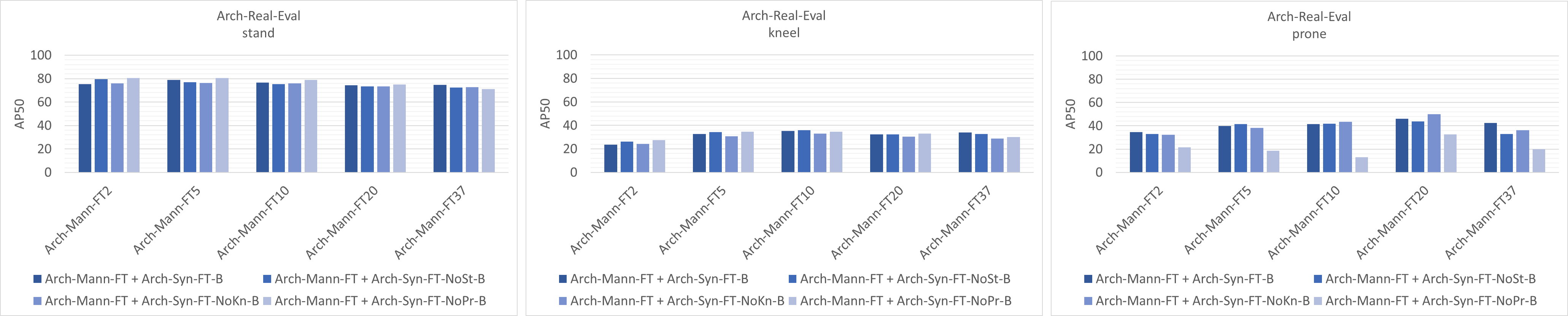}
\caption{AP50 of \textit{YOLOv5n6} fine-tuned on the various balanced hybrid sets constructed by the three leave-one-pose-out synthetic fine-tuning datasets (i.e., \textit{Arch-Syn-FT-NoSt-B}, \textit{Arch-Syn-FT-NoKn-B} and \textit{Arch-Syn-FT-NoPr-B}) and the different subsets of \textit{Arch-Mann-FT37}.}
\label{fig:ft_mann_syn_b_nano_pose}
\end{figure*}

The results are presented in Fig.~\ref{fig:ft_mann_syn_b_nano_pose}. In general, the detection accuracy of \textit{stand} and \textit{kneel} did not obviously change when we removed any one of the poses from the synthetic fine-tuning dataset. However, the performance of \textit{prone} degraded drastically when we excluded the virtual characters in \textit{prone} from the fine-tuning dataset, which strongly supports our earlier claim.\smallskip 

\begin{figure*}
\centering
\includegraphics[width=0.8\textwidth]{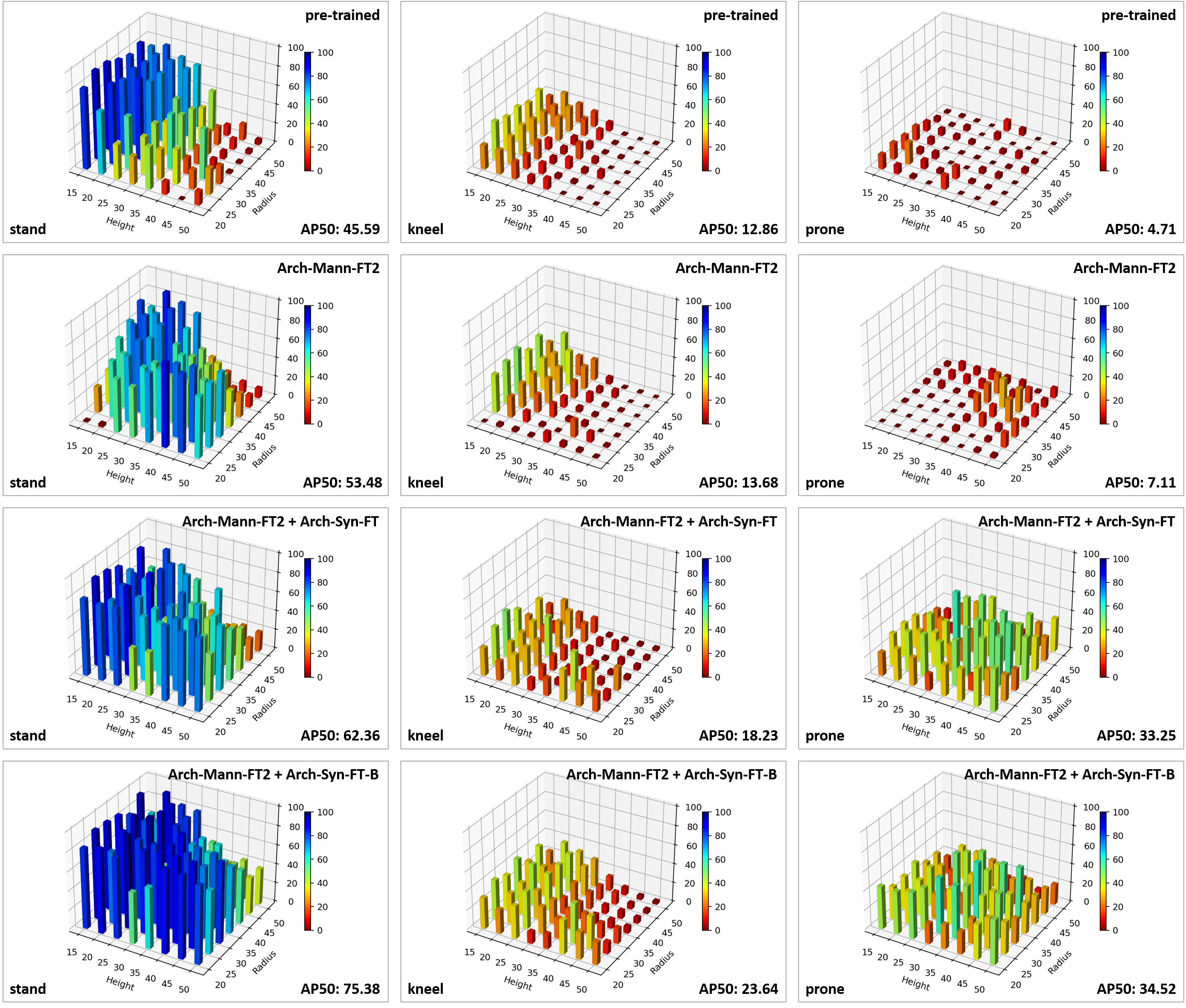}
\caption{AP50 comparison on the altitude/radius grid of \textit{YOLOv5n6} fine-tuned on the different hybrid sets of \textit{Arch-Mann-FT2}, \textit{Arch-Syn-FT} and \textit{Arch-Syn-FT-B}.}
\label{fig:ft_altitude_radius}
\end{figure*}

\noindent{\bf Precise Model Diagnosis: Performance Comparison on the Altitude/Radius Grid.} So far, we have demonstrated that the detection accuracy of the pre-trained YOLOv5 models on the UAV-based evaluation dataset can be considerably boosted by fine-tuning the models on the hybrid sets of UAV-based real and synthetic data. For example, we have shown that the AP50 of the pre-trained \textit{YOLOv5n6} can be increased by about 30 in AP value by fine-tuning it on a joint set of \textit{Arch-Mann-FT2} and \textit{Arch-Syn-FT-B} (Fig. \ref{fig:ft_mann_syn_b_nano}). In this section, we further analyze this particular example with the complete information about the UAV positions over the altitude/radius grid as shown in Fig.~\ref{fig:ft_altitude_radius}. Our goal is to give an idea of how to utilize the metadata provided by \textit{Archangel} to diagnose problems with a UAV-based object detection model.

The results are presented in Fig.~\ref{fig:ft_altitude_radius}. In this figure, we can clearly observe how the pre-trained \textit{YOLOv5n6}'s performance gradually progresses with the different fine-tuning datasets, from the real-data-only dataset to the unbalanced hybrid dataset to the balanced hybrid dataset. Initially, the pre-trained \textit{YOLOv5n6} performs fairly well at the low altitudes but fails at the high altitudes due to the curse of the pre-training dataset, which is composed mostly of ground-based human instances. After being fine-tuned on \textit{Arch-Mann-FT2}, the fine-tuned \textit{YOLOv5n6} gets much better at detecting human instances from a relatively higher altitude or larger circle radii. However, the detection accuracy of the human instances at close range decreases considerably. We argue that this is mainly because the image resolution of \textit{Arch-Mann-FT2} (i.e., 1920x1080) is much larger than that of the evaluation dataset (i.e., 1304x978) and the bounding boxes contained in \textit{Arch-Mann-FT2} are generally small. In other words, fine-tuning \textit{YOLOv5n6} on \textit{Arch-Mann-FT2} inevitably causes a bias on the model toward detecting tiny objects. In contrast, this negative effect does not occur when we fine-tune the model on the hybrid set of \textit{Arch-Mann-FT2} and \textit{Arch-Syn-FT} (or \textit{Arch-Syn-FT-B}). We believe that this is because our synthetic dataset covers an extended range of UAV positions over the grid so as to cause less bias on the fine-tuning process.

Based on the results shown in Fig.~\ref{fig:ft_altitude_radius}, to further improve the model's detection accuracy, we can focus on improving the model's performance at high altitudes and large circle radii, or the performance of non-standing positions, such as \textit{kneel} and \textit{prone}. Notably, we find that the performance improvement of \textit{kneel} is surprisingly small to high altitudes and large circle radii, which requires a deeper investigation in future work.

\section{Discussion and Future Direction}
\label{sec:discussion}
With the series of experiments presented in Sec.~\ref{sec:results}~and~\ref{sec:ablation}, we have demonstrated the distinctive value of the \textit{Archangel} dataset. Particularly, we have clearly illustrated how to utilize the dataset's metadata to evaluate and diagnose the UAV-based object detectors on the altitude/radius grid. Moreover, we have systematically analyzed how to involve both real and synthetic data within the UAV-based fine-tuning process. Such fundamental studies of UAV-based perception had not been achieved until we curated \textit{Archangel}.

Despite having all these merits, \textit{Archangel} is still in its early stage and has much room for improvement and development. In the following, we suggest a few possible future directions regarding \textit{Archangel}:\smallskip

\noindent{\bf Direct Extension of this Study.} Many results shown in Sec.~\ref{sec:results}~and~\ref{sec:ablation} imply that increasing the data diversity of \textit{Archangel} is one of the most promising future research directions. For instance, since we have demonstrated that fine-tuning models with virtual characters in unusual poses is particularly effective, we can include more atypical poses into \textit{Archangel} to further explore this phenomenon, which is crucial especially in search and rescue scenarios where finding people in severe physiological states is the priority. Additionally, we can diversify the appearances/attributes of either the real or synthetic human instances in the dataset, investigating whether this will resolve the issue of overfitting as we have discussed earlier. For the same purpose, we can increase the diversity of components beyond the foreground objects, such as the real and synthetic backgrounds included in \textit{Archangel}. Finally, we can extend \textit{Archangel} to include more object categories, such as various types of vehicles, which frequently exist in other UAV-based object detection datasets (Tab.~\ref{tab:dataset-summary}) so that \textit{Archangel} can be used in conjunction with those datasets.

Next, exploring more sophisticated fine-tuning strategies is another potential direction to extend this work. In this study, we have demonstrated that the performance of the pre-trained SoTA object detector can be boosted considerably by fine-tuning the model on a balanced UAV-based fine-tuning dataset constructed by directly merging a real subset and a synthetic subset. Nevertheless, it is worth exploring if there is a better strategy for sampling each subset or merging the two subsets. For instance, to build up a balanced fine-tuning dataset, instead of randomly selected samples from the synthetic fine-tuning dataset, we may do the sampling based on certain distance measurements.\smallskip

\noindent{\bf UAV-based Visual Representation Learning with Metadata.} In this study, we used the position and pose metadata provided by \textit{Archangel} only for accurate model evaluation and diagnosis. However, we believe that there is a huge potential to utilize such metadata during training for better visual representation learning. A notable example for this is NDFT \cite{NDFT}, where the authors exploited adversarial training with the coarse metadata labeled by themselves to enhance the robustness of the learned features for UAV-based object detection. We expect that such a framework will benefit greatly from the extensive metadata provided by \textit{Archangel}. Beyond UAV-based perception, in medical \cite{vu2021medaug} and underwater \cite{yamada2021leveraging} imaging, it has also been shown that dataset metadata is useful for learning visual representation with self-supervised learning or contrastive learning.\smallskip

\noindent{\bf UAV-based Synthetic Data Generation and Augmentation.} We have found that fine-tuning larger models with the synthetic data that we generated often causes the issue of overfitting. This issue might be mitigated by directly synthesizing more diverse images. However, as we have discussed, there is usually a huge domain gap between the synthetic data and real data in terms of object appearances/attributes and scene structures, which may not be solved by simply increasing the number of synthetic images. Hence, addressing this domain gap issue within the scope of UAV-based perception is another important future research direction for \textit{Archangel}. Possible solutions include: (1) jointly training an object detector with a generative model which transforms synthetic images to be more visually realistic \cite{liu2019generative, shen2023progressive}, and (2) formulating the process of synthetic data generation as a learning problem to synthesize scene structures better matching real-world scene distributions \cite{kar2019meta}.

\section{Conclusion}
In this paper, we introduce a unique UAV-based object detection dataset, \textit{Archangel}, to encourage the community to continue developing more effective UAV-based object detection approaches with dataset metadata and synthetic data. A comprehensive study is carefully designed to show how to utilize \textit{Archangel} to fully optimize a state-of-the-art object detector with a hybrid fine-tuning dataset comprising both real and synthetic data. Additionally, we also demonstrate the huge benefit of leveraging the dataset metadata during model evaluation by comparing the performance of the model across the different object poses and UAV positions. As we have discussed, although there is still much room for improvement, we hope that \textit{Archangel} is useful for the broader machine learning community and can lead to future advances in the area of UAV-based perception.

\section*{Acknowledgment}
This research was sponsored by the Defense Threat Reduction Agency (DTRA) and the DEVCOM Army Research Laboratory (ARL) under Grant No. W911NF2120076. The authors would like to thank the US Army Artificial Innovation Institute (A2I2) for supporting the annotation of \textit{Archangel}.

\bibliographystyle{unsrt}
\bibliography{refs}

\begin{IEEEbiography}[{\includegraphics[width=1in,height=1.25in,clip,keepaspectratio]{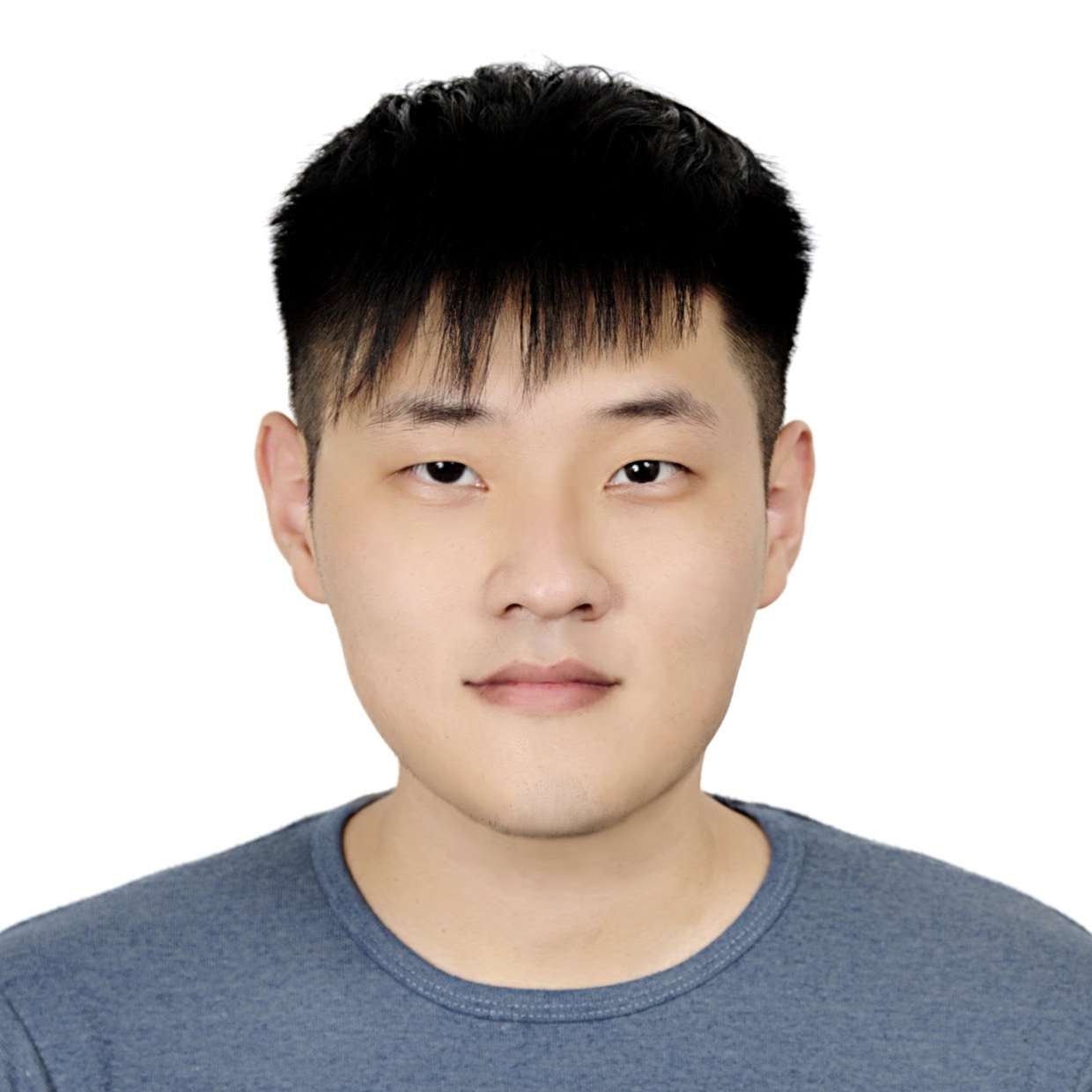}}]{Yi-Ting Shen}
received the B.S. degree in Electrical Engineering from National Taiwan University in 2016 and the M.S. degree in Electronics Engineering from National Taiwan University in 2019. He joined the Department of Electrical and Computer Engineering at the University of Maryland, College Park as a Ph.D. student in 2020. His research interests include computer vision, machine learning and hardware/software co-design.   
\end{IEEEbiography}

\begin{IEEEbiography}[{\includegraphics[width=1in,height=1.25in,clip,keepaspectratio]{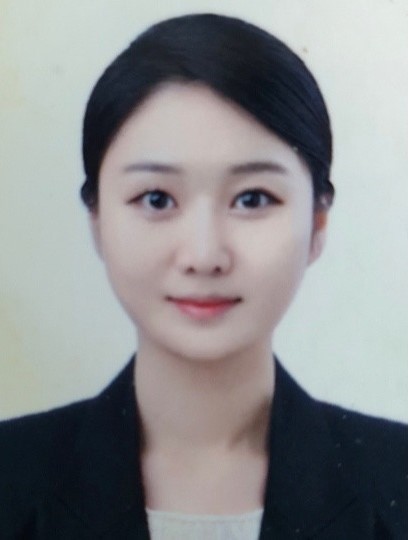}}]{Yaesop Lee}
received the bachelor’s degree in Electrical Engineering from Sogang University, South Korea, and master’s degree in ENTS from University of Maryland, College Park. She is currently pursuing Ph.D. degree in the Department of Electrical and Computer Engineering at the University of Maryland, College Park. Her research interests include Embedded Computer Vision and Real-time Image/Signal Processing.    
\end{IEEEbiography}

\begin{IEEEbiography}[{\includegraphics[width=1in,height=1.25in,clip,keepaspectratio]{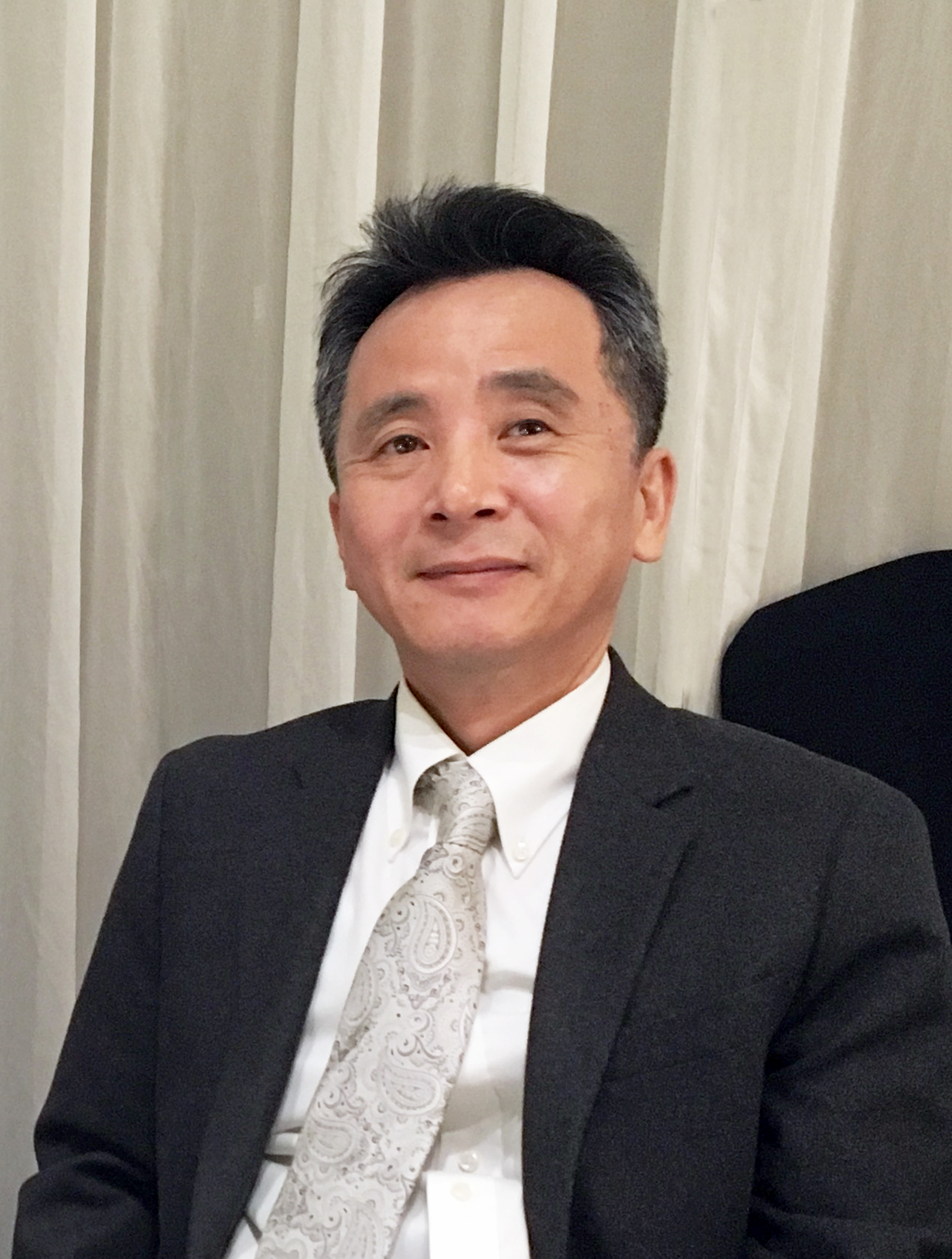}}]{Heesung Kwon}
is Senior Researcher and Team Lead at the DEVCOM Army Research Laboratory (ARL). He received the B.Sc. degree in Electronic Engineering from Sogang University, Seoul, Korea, in 1984, and the MS and Ph.D. degrees in Electrical Engineering from the State University of New York at Buffalo in 1995 and 1999, respectively. 

Dr. Kwon served as one of the government leads of an ARL collaborative research program--the Internet of Battlefield Things (IoBT). Dr. Kwon also served as Associate Editor of IEEE Trans. on Aerospace and Electronic Systems. He has published over 150 journal papers, book chapters, and conference papers on various topics. Dr. Kwon is a co-recipient of the best paper award at the Army Science Conference in 2004 and the best paper runner-up award at the IEEE International Conference on Biometrics: Theory, Applications, and Systems (BTAS 2016). He has been on Technical Program Committee for various conferences and workshops relevant to image/video analytics and machine learning.
\end{IEEEbiography}

\begin{IEEEbiography}[{\includegraphics[width=1in,height=1.25in,clip,keepaspectratio]{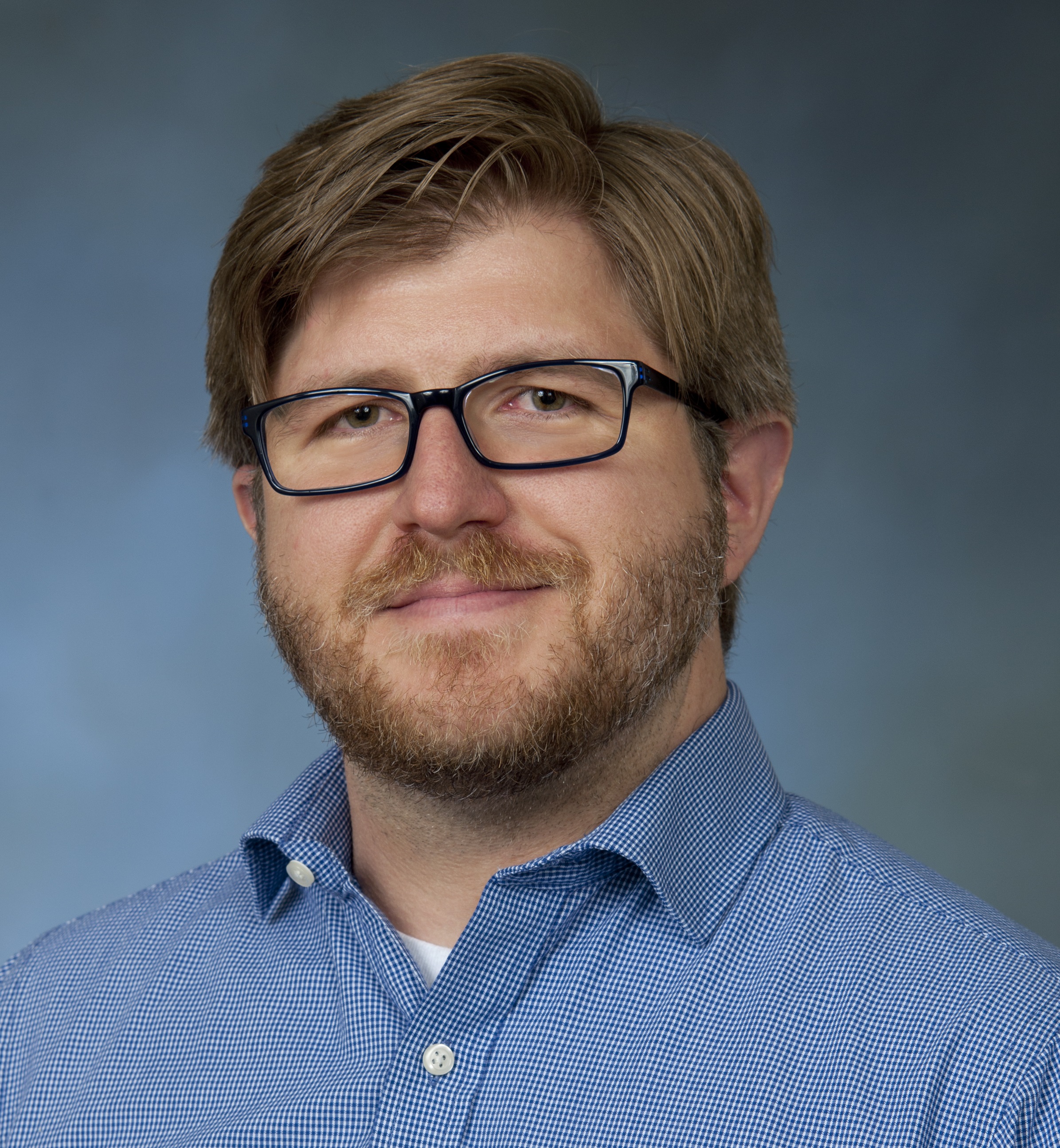}}]{Damon M. Conover}
is an electtrical engineer in the Intelligent Perception Branch at the DEVCOM Army Research Laboratory (ARL). His research focuses on 3D geospatial data processing and visualization, simulation and synthetic data generation, and robotic systems. He is particularly interested in the intersection of geospatial information and robotics for high-level planning. Dr. Conover earned his Ph.D. from George Washington University in 2015.
\end{IEEEbiography}

\begin{IEEEbiography}[{\includegraphics[width=1in,height=1.25in,clip,keepaspectratio]{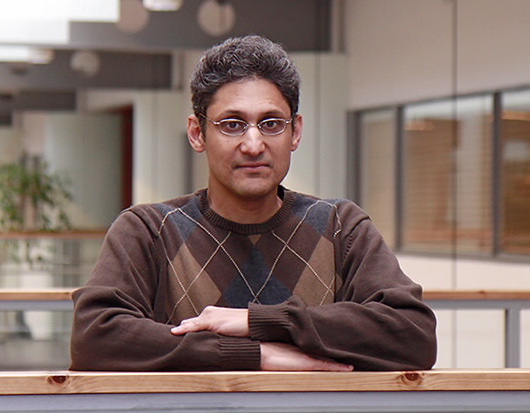}}]{Shuvra S. Bhattacharyya}
is a Professor in the Department of Electrical and Computer Engineering at the University of Maryland, College Park. He holds a joint appointment in the University of Maryland Institute for Advanced Computer Studies (UMIACS), and is affiliated with the Maryland Crime Research and Innovation Center (MCRIC). He also holds a part-time visiting position as Chair of Excellence in Design Methodologies and Tools at Institut National Des Sciences Appliquées (INSA) in Rennes, France. He received the Ph.D. degree from the University of California at Berkeley. He has held industrial positions as a Researcher at the Hitachi America Semiconductor Research Laboratory, and Compiler Developer at Kuck \& Associates. From 2015 through 2018, he was a part-time visiting professor in the Department of Pervasive Computing at the Tampere University of Technology (now Tampere University), Finland, as part of the Finland Distinguished Professor Programme. He has also held visiting research positions with the U.S. Air Force Research Laboratory (AFRL). He is a Fellow of the IEEE.
\end{IEEEbiography}

\begin{IEEEbiography}[{\includegraphics[width=1in,height=1.25in,clip,keepaspectratio]{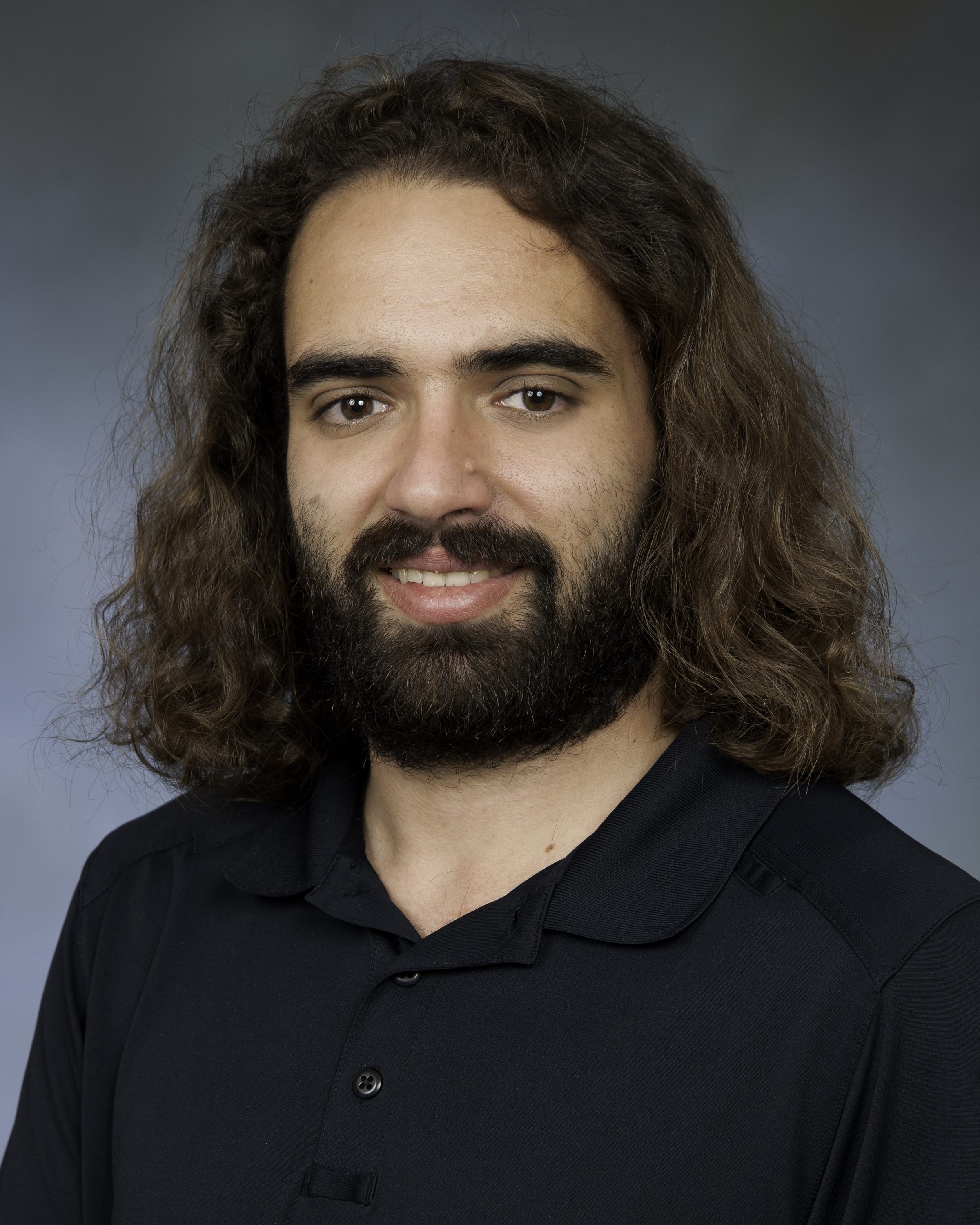}}]{Nikolas Vale}
is a mechanical engineer at the DEVCOM Army Research Laboratory (ARL) under the Autonomous Sensing \& Integration Branch. His focuses are on designing, constructing, programming and operating unmanned aerial vehicle (UAV) research platforms. He is particularly interested in developing modular, reconfigurable hardware and software systems for the purposes of data collection, experimentation and prototyping.
\end{IEEEbiography}

\begin{IEEEbiography}[{\includegraphics[width=1in,height=1.25in,clip,keepaspectratio]{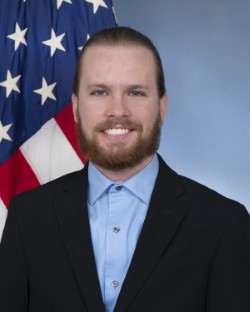}}]{Joshua D. Gray}
is a mechanical engineer at the DEVCOM Army Research Laboratory (ARL) under the Autonomous Sensing \& Integration Branch. His focuses are on designing, constructing, and operating unmanned aerial vehicle (UAV) research platforms. He is particularly interested in developing small, multi-rotor UAVs for experimentation and data collection.
\end{IEEEbiography}

\begin{IEEEbiography}[{\includegraphics[width=1in,height=1.25in,clip,keepaspectratio]{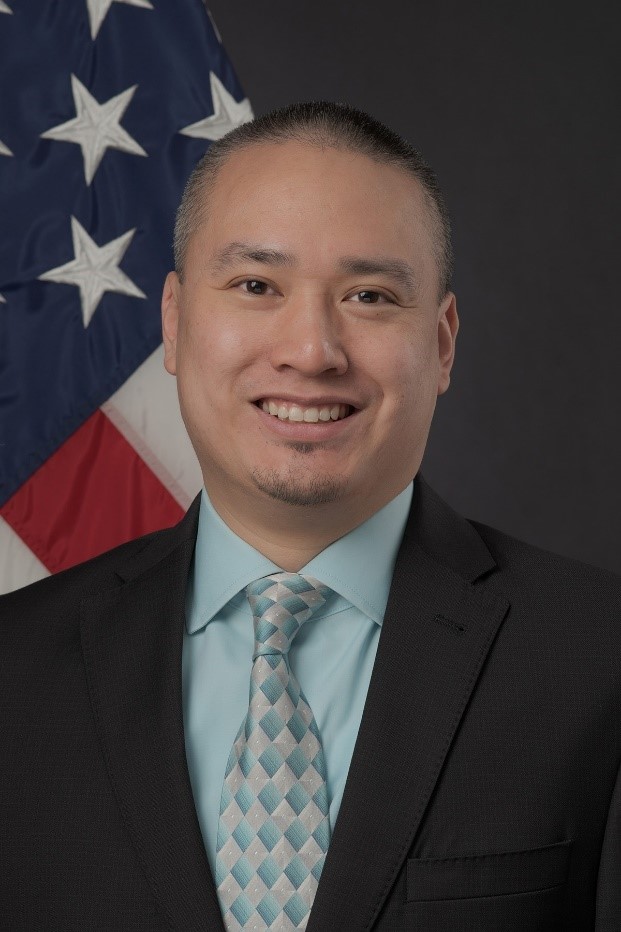}}]{G. Jeremy Leong}
serves as a Program Manager within DTRA’s Counter-WMD Advanced Research Division. Prior to joining DTRA, Jeremy spent 13 years in energy research; most recently 6 years at the U.S. Department of Energy where he oversaw several programs in fuels and chemicals, advanced materials, as well as modeling \& simulation/high performance computing for multiple directorates. In the past 8 years, he managed a total Federal applied R\&D budget of over \$180M with numerous recognitions for successful technology transitions and deployments. Dr. Leong earned his Ph.D. in Applied Chemistry (Catalyst Science and Engineering) at the Colorado School of Mines. Since 2009, he has authored more than 30 peer reviewed manuscripts, internal assessments, and congressional reports. He has also served on numerous technical panels and presented more than two dozen conference talks on modeling \& simulation, software development, as well as the chemistry of materials and their practical applications.
\end{IEEEbiography}

\begin{IEEEbiography}[{\includegraphics[width=1in,height=1.25in,clip,keepaspectratio]{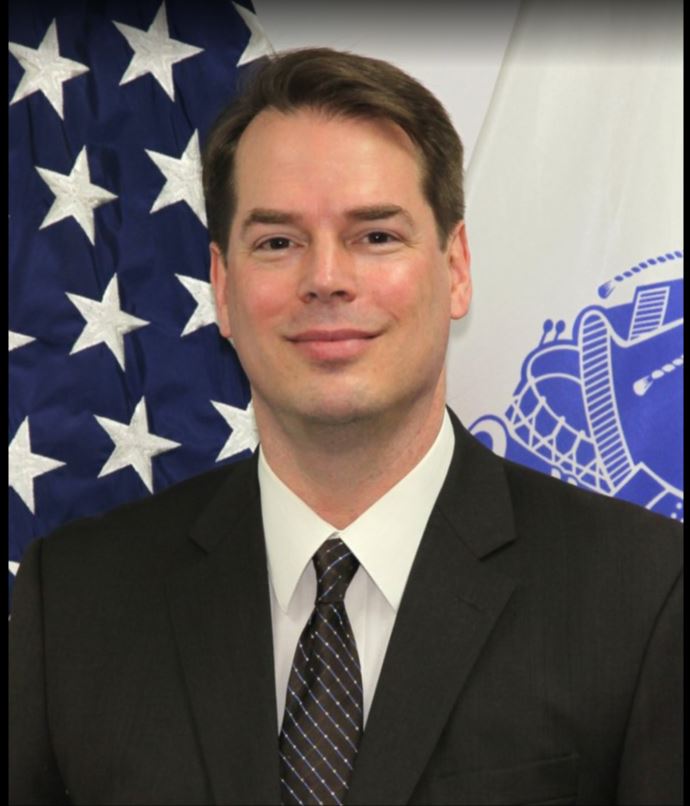}}]{Kenneth Evensen}
served twenty-seven years in various levels of command in the United States Army. Notably he served as the Program Manager for developing the next generation of tactical networking for the Department of Defense within the Joint Tactical Radio System Program Executive Office. He retired as an active member of the Army Acquisition Corps and upon retirement he named as the Director of the United States Army International Technology Center - Pacific from late 2012 thru early 2018 and was responsible for technology search and coordinating cooperation opportunities in science and technology throughout the Pacific Region on behalf of the US Army. He currently serves as the Division Chief for the Advanced Research Division within the Counter WMD Technologies Department, Research \& Development Directorate, Defense Threat Reduction Agency. 
\end{IEEEbiography}

\begin{IEEEbiography}[{\includegraphics[width=1in,height=1.25in,clip,keepaspectratio]{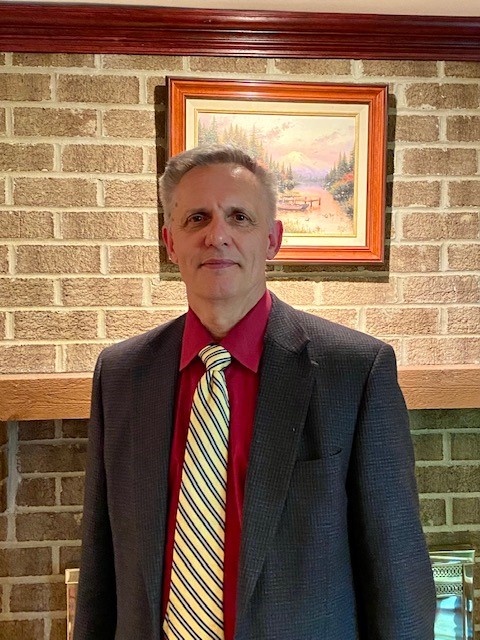}}]{Frank Skirlo}
presently supports DTRA’s Advanced Research Division, Counter-WMD Department, Research \& Development Directorate as a Technical Program Analyst. He served as a U.S. Army Signal Officer for over 24 years of active service, and retired at the rank of Colonel in December 2011, after completing a variety of worldwide assignments, including command and staff positions in Germany, Korea, and Hawaii. His last assignment was with the Joint Staff J-6, overseeing C4 transport programs and requirements. After retiring from the military, Dr. Skirlo supported the U.S. Army C5ISR Center’s Night Vision and Electronics Sensors (now Research and Technology Integration) Directorate’s Integrated Sensor Architecture (ISA) project. Dr. Skirlo earned his Ph.D. in Electrical Engineering from George Mason University, with his thesis focusing on image processing and UAS-based aided target detection. He also has a Master of Strategic Studies from the U.S. Army War College, Master of Science degree in Electrical Engineering from the University of Colorado at Colorado Springs, and a Bachelor of Science degree from the University of Florida. 
\end{IEEEbiography}

\EOD

\end{document}